\documentclass{article}

\usepackage{arxiv}
\usepackage{pgfplots}
\pgfplotsset{compat=1.18}
\usepackage[utf8]{inputenc} 
\usepackage[T1]{fontenc}    
\usepackage{hyperref}       
\usepackage{url}            
\usepackage{booktabs}       
\usepackage{amsmath,amsfonts,amssymb}       
\usepackage{nicefrac}       
\usepackage{microtype}      
\usepackage{cleveref}       
\usepackage[noend]{algorithmic}
\usepackage{algorithm}
\usepackage[authoryear]{natbib}
\usepackage{xurl}
\usepackage{tikz}
\usepackage{multirow}
\usepackage{graphicx}
\usepackage[nolist]{acronym}

\usepackage[group-separator={,},group-minimum-digits=4]{siunitx}
\usepackage{makecell}
\usepackage{chemmacros}
\definecolor{Unchanged}{RGB}{67,1,84}
\definecolor{NewBuild}{RGB}{0,183,255}
\definecolor{Demol}{RGB}{0,12,235}
\definecolor{VegeN}{RGB}{0,217,33}
\definecolor{VegeG}{RGB}{255,230,0}
\definecolor{VegeR}{RGB}{255,140,0}
\definecolor{MOch}{RGB}{255,0,0}

\definecolor{Ground}{RGB}{0,21,181}
\definecolor{Building}{RGB}{51,139,255}
\definecolor{Vegetation}{RGB}{84,219,0}
\definecolor{MO}{RGB}{255,196,0}
\usepackage{natbib}
\usepackage{doi}

\title{DC3DCD: unsupervised learning for multiclass 3D point cloud change detection}

\date{}

\usepackage{authblk}

\author[1,2\thanks{\tt{iris.de-gelis@irisa.fr}}]{Iris de~G\'{e}lis}
\author[2]{S\'{e}bastien Lef\`{e}vre}
\author[3]{Thomas Corpetti}

\affil[1]{Magellium, Toulouse, France}
\affil[2]{IRISA, UMR 6074, Universit\'{e} Bretagne Sud, Vannes, France}
\affil[3]{CNRS, LETG, UMR 6554, Rennes, France}

\hypersetup{
pdftitle={DC3DCD: unsupervised learning for multiclass 3D point cloud change detection},
pdfauthor={Iris de~G\'{e}lis, S\'{e}bastien Lef\`{e}vre, Thomas Corpetti},
pdfkeywords={3D point clouds, Change detection, Unsupervised Deep learning, Deep clustering},
}

\begin{document}
\maketitle

\begin{abstract}
	In a constant evolving world, change detection is of prime importance to keep updated maps. To better sense areas with complex geometry (urban areas in particular), considering 3D data appears to be an interesting alternative to classical 2D images. In this context, 3D point clouds (PCs), whether obtained through LiDAR or photogrammetric techniques, provide valuable information. While recent studies showed the considerable benefit of using deep learning-based methods to detect and characterize changes into raw 3D PCs, these studies rely on large annotated training data to obtain accurate results. The collection of these annotations are tricky and time-consuming. The availability of unsupervised or weakly supervised approaches is then of prime interest. In this paper, we propose an unsupervised method, called DeepCluster 3D Change Detection (DC3DCD), to detect and categorize multiclass changes at point level. We classify our approach in the unsupervised family given the fact that we extract in a completely unsupervised way a number of clusters associated with potential changes. Let us precise that in the end of the process, the user has only to assign a label to each of these clusters to derive the final change map. Our method builds upon the DeepCluster approach, originally designed for image classification, to handle complex raw 3D PCs and perform change segmentation task. An assessment of the method on both simulated and real public dataset is provided. The proposed method allows to outperform fully-supervised traditional machine learning algorithm and to be competitive with  fully-supervised deep learning networks applied on rasterization of 3D PCs with a mean of IoU over classes of change of 57.06\% and 66.69\% for the simulated and the real datasets, respectively. The code is available at \url{https://github.com/IdeGelis/torch-points3d-DC3DCD}.
\end{abstract}


\section{Introduction}\label{sec:intro}
While urban environments are continuously and rapidly evolving, change detection between several temporal acquisitions is a way to quickly highlight modified areas in order to update maps \citep{demir2012updating} or to identify damaged objects \citep{dong2013comprehensive} in case of natural disasters. With regard to the complexity of urban landscapes, sensing the area using also vertical information appears to be judicious. Indeed, such  \ac{3D} information allows to better characterize environment geometry and to avoid \ac{2D} image problems such as the difference of viewing angles between distinct acquisitions, spectral variability of objects over time, perspective, and distortion effects \citep{qin20163d}. \ac{3D} data are acquired thanks to \ac{LiDAR} sensor or photogrammetric process, both resulting in \ac{3D} \acp{PC}. No matter the type of acquisition, multi-temporal \ac{3D} data are generalizing. Indeed, more and more national mapping agencies opt for full territory \ac{ALS} campaign, as in the Netherlands with \ac{AHN} multi-temporal campaigns \citep{sande2010assessment}, or in France with the \ac{LiDAR} \ac{HD} project whose objective is to propose a complete 3D high resolution coverage of France with regular updates in the future. In the same time satellites mission for \ac{3D} sensing are multiplying, e.g., Pléiades \citep{bernard20123d}, Pléiades Néo \citep{jerome2019shaping}, \ac{CO3D} \citep{lebegue2020co3d} missions. In civil engineering, \ac{TLS} or \ac{UAV} photogrammetry are becoming unavoidable for an accurate sense of complex objects. Thereby, this calls for methods able to analyze these multi-temporal \ac{3D} data.

Whether related to urban environment \citep{stilla2023change} or geosciences \citep{okyay2019airborne}, many studies have been focused on handling 3D data for accurate change extraction. Recently, some methods based on deep learning proved their efficiency over traditional distance-based methods and machine learning. While the first deep learning methods for 3D change detection were based on 2.5D rasterization of PCs into \ac{DSM} \citep{zhang2018change, zhang2019detecting} or range image \citep{nagy2021changegan}, most recent works are dealing directly with the raw 3D data. Indeed, although easing the computation, any rasterization process implies a significant loss of information for instance on building facades but also due to aggregation of several points in a cell.  \cite{KU2021192} propose to represent \acp{PC} by graphs and apply graph convolution operator (EdgeConv) \citep{wang2019dynamic} to extract discriminative features. However, their proposed method, called \ac{SiamGCN}, results only in change detection at the scene level, i.e., solving a change classification task. On the opposite, \cite{degelis2023siamese} proposed a network to solve change segmentation task, i.e., providing multiclass change information at point level (so coarser results than change classification). To handle raw \ac{3D} \acp{PC}, they rely on \ac{KPConv} \citep{thomas2019kpconv}. Their network, named Siamese \ac{KPConv} enables to outperform other machine learning methods or \ac{DSM}-based deep learning methods. More recently, a study \citep{degelis2023change} suggests improving Siamese \ac{KPConv} by making the network focusing more on change-related features. To do so, \cite{degelis2023change} proposes to provide a hand-crafted feature related to change as input to Siamese \ac{KPConv} along with \ac{3D} point coordinates. They also developed three other architectures for this particular task. OneConvFusion and Triplet \ac{KPConv} contain specific change-related encoder that takes as input only feature difference. Encoder Fusion SiamKPConv (EFSKPConv) concatenates both change and mono-date information directly in the encoder. This latter achieves the best results compared to previous state-of-the-art methods.

Although efficient, these deep learning-based methods are supervised, i.e.,  require large databases for the training of the network. For change segmentation, millions of points should be annotated according to the change type. This annotation is often performed manually because any automatization process is not obvious due to \acp{PC} characteristics such as the lack of point-to-point matching, sparsity, or occlusions (see the example of \ac{AHN-CD} dataset \citep{degelis2023siamese}). Therefore, it is important to develop methods that require none or at least less annotations to perform \ac{3D} change segmentation. As stated in recent surveys \citep{kharroubi2022three,xiao20233d}, the literature still lacks of methods for unsupervised or weakly supervised learning when dealing with \ac{3D} \acp{PC} change detection. Indeed, nowadays unsupervised methods are mostly traditional rule-based methods that are often very specific to a dataset. Also, recently an adaptation of \ac{DCVA} \citep{saha2019unsupervised} to \ac{3D} \acp{PC} change detection has been proposed based on self-supervised learning \citep{degelis2023deep}. However, this unsupervised method only deals with binary change segmentation, thus the method is not able to distinguish between multiple classes of changes. Thereby, in this paper, we propose an unsupervised learning strategy to deal with multiclass \ac{3D} \acp{PC} change segmentation. The strategy is based on deep clustering principle and in particular on DeepCluster \citep{caron2018deep}. Deep clustering consists in jointly optimizing deep representation of the data and performing clustering with learned features \citep{ren2022deep,zhou2022comprehensive}. This strategy has received increasing interest for \ac{2D} image unsupervised representation learning in computer vision \citep{caron2018deep,cho2021picie}. However, as far as the change detection task is concerned, the use of deep clustering is less common. \cite{zhang2018unsupervised} and \cite{dong2021multiscale} propose to rely on an unsupervised clustering of deep feature representations to further train their network to perform change detection. In \cite{zhang2018unsupervised} a stack of fully-connected layers is used to learn Gaussian-distributed and discriminative difference representations for non-change and different types of changes. \cite{dong2021multiscale} further improves the latter by using a \ac{CNN} relying on multi-scale self-attention. Another study partially uses deep clustering principle for \ac{2D} binary change detection by introducing a deep clustering loss jointly with contrastive and appealing losses to make a \ac{CNN} network learning discriminative mono-date features. These features are further compared using \ac{DCVA} principle to extract binary changes in multi-modal optical and \ac{SAR} images \citep{saha2021self}. 
This principle was adapted in \cite{degelis2023deep} for \ac{3D} \acp{PC} binary change detection.  Finally, to the best of our knowledge, \cite{zhang2021unsupervised} is the first study using the deep clustering principle on \ac{3D} data. However, the approach relies on a voxelisation of the \ac{PC} instead of dealing with the raw \ac{PC} directly, and performs classification at the scene level and not at the point level (a.k.a. segmentation).

The contributions of this work are thus as follows:
\begin{enumerate}
    \item To the best of our knowledge, we propose the first unsupervised learning strategy for multiclass change segmentation into raw \ac{3D} \acp{PC}.
    \item We build upon DeepCluster \citep{caron2018deep} to tackle the change segmentation task in \ac{3D} \acp{PC}. To do so, we experiment our model with several \ac{3D} \acp{PC} change segmentation architectures.
    \item After analyzing the learning behavior, we evaluate our method on both real and simulated public datasets, and provide comparisons with supervised, unsuperpervised and weakly-supervised methods.
    \item We demonstrate that our model, if provided with a perfect binary change map, can compete with fully supervised deep methods for multiclass change segmentation.
\end{enumerate}
The description of our method is given in Section~\ref{sec:method}. The results are provided in Section~\ref{sec:res} and discussed in Section~\ref{sec:discu}. Finally, the conclusion is provided in Section~\ref{sec:ccl}.
The implementation of our method will be made available at \textit{link}\footnote{The code will be made available upon publication.}.

\section{Unsupervised 3D change detection}\label{sec:method}
We describe here our 2-step methodology to compare two \ac{3D} \acp{PC} and classify the second one with several types of changes. Since our study does not focus on \acp{PC} registration, we assume that the \acp{PC} have been already registered. Various registration techniques exist and one can use for instance the Iterative Closest Point algorithm \citep{besl1992method}. The first step, called DC3DCD, is fully unsupervised and does not need any labeling of the \acp{PC}. It relies on the concept of deep clustering, that is first recalled in Sec.~\ref{sec:deepcluster}. We then explain how deep clustering can be used to solve the \ac{3D} \ac{PC} change detection problem in Sec.~\ref{subsec:DC3DCD}, where we detail the backbone model, the use of a prototype layer, the input data needed to feed the network, and some specific elements of the training procedure. This fully unsupervised step based on DC3DCD leads to a clustering of the \acp{PC} into different clusters or pseudo-labels. A second step is then required to map these pseudo-labels into real labels. Only here is the user involved, through a manual mapping process limited to the number of clusters (and not the number of points). This weakly-supervised step is detailed in Sec.~\ref{sec:DC3DCDmapping}.

\subsection{DeepCluster principle}\label{sec:deepcluster}

Among the variety of studies related to deep clustering \citep{ren2022deep, zhou2022comprehensive}, DeepCluster appears to be among the most fundamental ones. Proposed by \cite{caron2018deep}, this method resides in a rather simple idea of alternatively clustering deep latent representation of data to obtain pseudo-labels further used to train a \ac{CNN}. In particular, the convolutional network is trained in a supervised manner using pseudo-labels as objective for prediction. In a traditional supervised approach, giving a set of $N$ images $x_{n}$ ($n \in [1,N]$), a parametrized classifier $g_{W}$ predicts the correct labels ($y_{n}$) using the features extracted by $f_{\theta}(x_{n})$ ($W$ and $\theta$ being the parameters from the classifier and the back-bone convolutional model, respectively). They are optimized according to the following problem:
\begin{equation}\label{eq:paramOpt}
    \min_{\theta, W} \frac{1}{N} \sum^{N}_{n=1}\ell(g_{W}(f_{\theta}(x_{n})), y_{n})
\end{equation}
where $\ell$ is the loss function, a classical \ac{NLL} in their method. This cost function is minimized using standard mini-batch stochastic gradient descent and backpropagation to compute the gradient. The difficulty in an unsupervised setting is therefore to define $y_{n}$.

In DeepCluster, \cite{caron2018deep} proposed to rely on a classical clustering algorithm such as $k$-means \citep{macqueen1967classification} or \ac{PIC} \citep{lin2010power} to obtain a pseudo-label ($y_{PL_{n}}$) that is used instead of $y_{n}$. \cite{caron2018deep} showed that the choice of the clustering algorithm is not crucial. Thereby, for illustration purposes, we continue with the example of $k$-means algorithm. This clustering method matches data to $k$ groups (pseudo-clusters) by minimizing distance between each data and its corresponding cluster center, called centroid (and contained in the centroid matrix $C$ in practice).

Finally, the unsupervised training process alternates between i) clustering the output features of the back-bone convolutional model ($f_{\theta}(x_{n})$) (clustering step), and ii) updating parameters $\theta$ and $W$ using the obtained pseudo-labels ($y_{PL_{n}}$) thanks to Equation~\ref{eq:paramOpt} (training step). This relies on the fact that a \ac{MLP} classifier on top of a standard \ac{CNN} with randomly initialized weights ($\theta$) provides results far above from the chance (i.e., random) level \citep{noroozi2016unsupervised}.

In practice, a few tricks are required to prevent trivial solutions, e.g., assigning all the inputs to the same cluster. First, empty clusters are avoided by randomly dividing in two groups the largest cluster when an empty cluster appears. Second, if the pseudo-cluster representation of data is largely imbalanced, the deep model will tend to assign all data to the most represented pseudo-cluster. To counter this, they propose to sample input images during the training based on a uniform distribution over the pseudo-labels.


They showed the robustness of their method by training different architectures (Alexnet \citep{krizhevsky2017imagenet} and VGG-16 \citep{simonyan2014very}) on ImageNet \citep{5206848} or YFCC100M \citep{thomee2016yfcc100m} images datasets.

In the following, we adapt this principle to \ac{3D} \acp{PC} change detection.

\subsection{DC3DCD: unsupervised learning for 3D multiple change extraction}\label{subsec:DC3DCD}
\begin{figure*}
    \centering
    \includegraphics[width=0.8\textwidth]{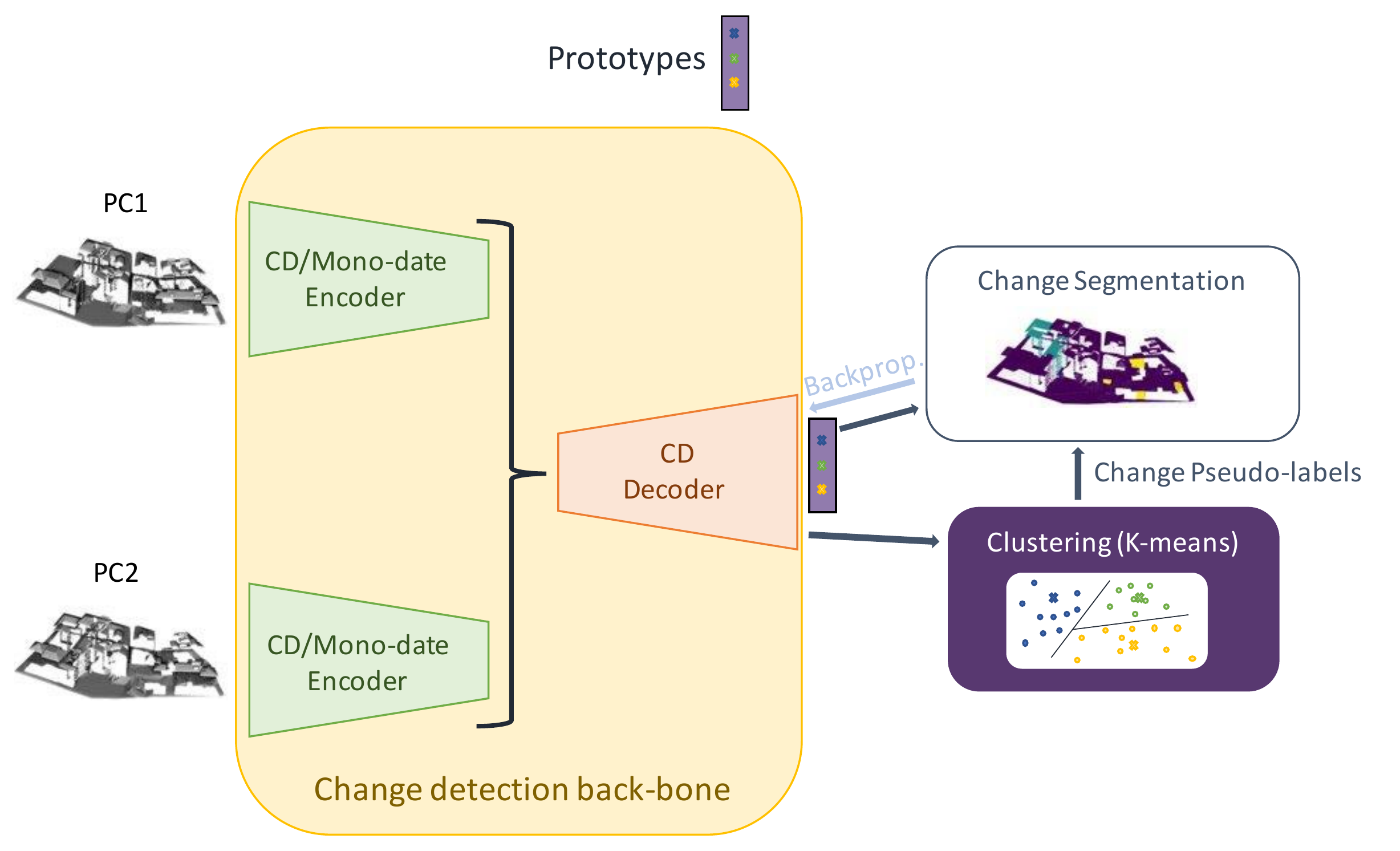}
    \caption[Illustration of our proposed method: DC3DCD]{\textbf{Illustration of our proposed method: DC3DCD.} It is trained by alternatively clustering deep features to match a pseudo-label to each point of PC 2. These pseudo-labels are used to optimize the back-bone trainable parameters.}
    \label{fig:dc3dcd}
\end{figure*}

Whereas the task and the data (\ac{2D} image classification) of DeepCluster is far from \ac{3D} \acp{PC} multiple change segmentation, we nevertheless decided to adapt this method to our task and particular data. By replacing the \ac{CNN} by a \ac{3D} \acp{PC} change detection back-bone, some change-related features can be extracted.
Thereby, the clustering of these deep features results in change-related pseudo-clusters. We further rely on these pseudo-clusters to optimize the trainable parameters $\theta$ of the change detection back-bone. Figure~\ref{fig:dc3dcd} illustrates our method called \ac{DC3DCD}.

\begin{algorithm}
\caption{\small Fully unsupervised DeepCluster 3DCD training}\label{alg:DC3DCD}
\begin{algorithmic}
\scriptsize{
\STATE Initialize the back-bone trainable parameters $\theta$
\FOR{$e \gets 1$ to $\mathcal{E}$}
\STATE Run mini-batch $k$-means to obtain centroids $C$ on the whole training set
\STATE Assign to each point of the training set a pseudo-cluster
\STATE Replace parameters of the prototype layer by $C$
\STATE Compute the weights $W_{k}$ considering pseudo-label distribution in the training set
\STATE Training sample selection: random drawing considering $W_{k}$
\FOR{$i \gets 1$ to $\mathcal{I}$}
\STATE Use $\mathcal{L}_{NLL}$ (weighted by $W_{k}$) to modulate the back-bone trainable parameters ($\theta$) considering pseudo-labels
\ENDFOR
\ENDFOR
}
\end{algorithmic}
\end{algorithm}

The overall training process of our method is given in the Algorithm~\ref{alg:DC3DCD}. Even if the general idea of DeepCluster remains, some specific features of \ac{DC3DCD} distinguishing it from the original DeepCluster idea should be noted, as described in the following.
    \subsubsection{Back-bone model} 
\begin{figure}
        \centering
         \centering
    \begin{minipage}[t]{0.48\textwidth}
  \centering
  \centerline{\includegraphics[width=\textwidth]{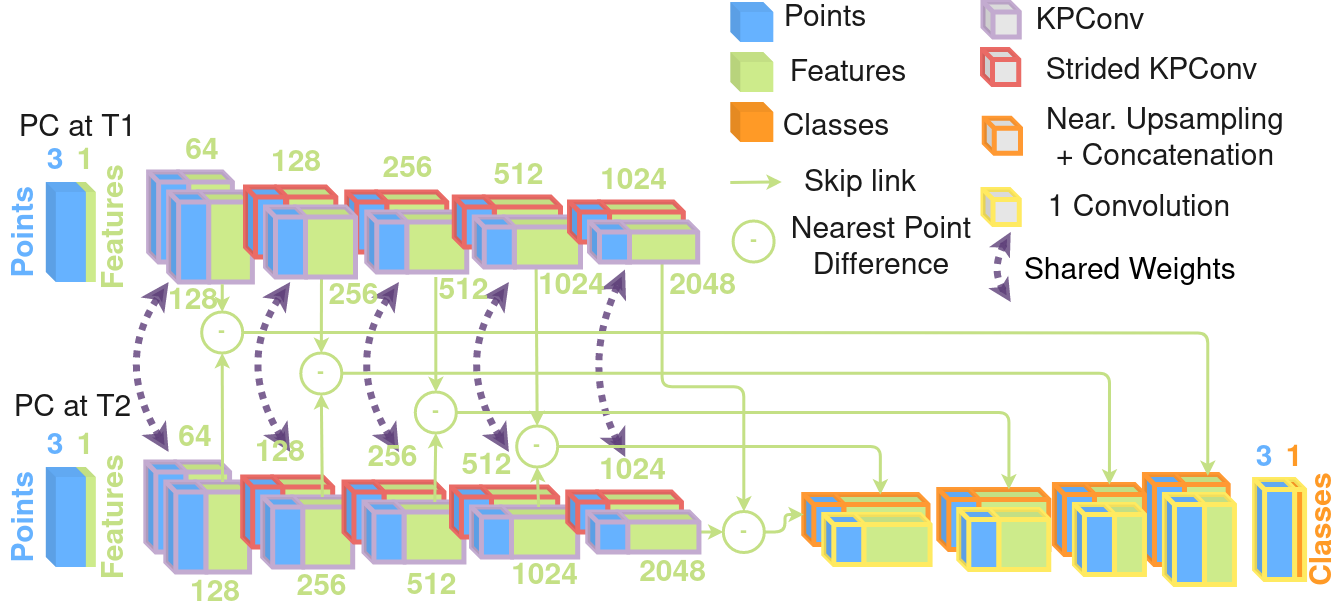}}
  \centerline{(\textbf{a}) Siamese KPConv \citep{degelis2023siamese} }\medskip
    \end{minipage}
    \begin{minipage}[t]{0.48\textwidth}
  \centering
  \centerline{\includegraphics[width=\textwidth]{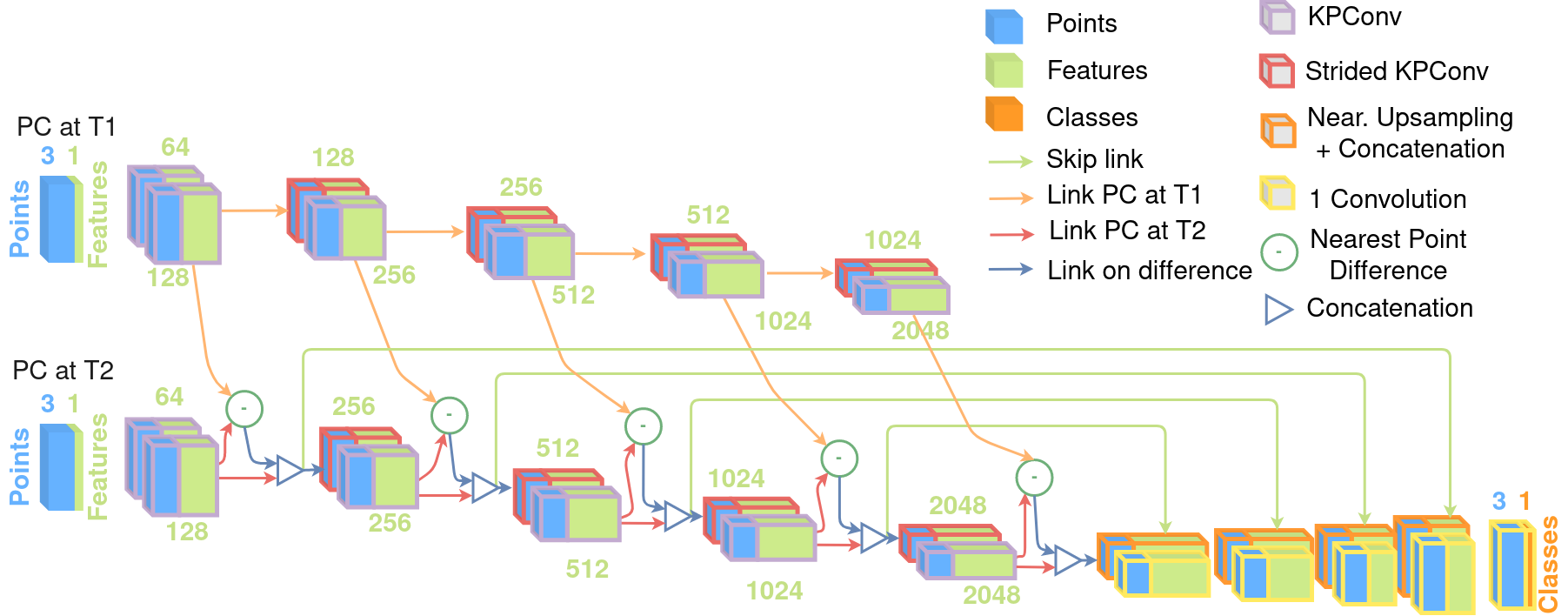}}
  \centerline{(\textbf{b}) Encoder Fusion SiamKPConv \citep{degelis2023change} }\medskip
    \end{minipage}
        \caption{\textbf{Back-bone architectures used in our experiments.}}
        \label{fig:backbones}
    \end{figure}

Because of the unstructured nature of \ac{3D} \acp{PC}, traditional \ac{CNN} models cannot be applied on them and furthermore, they do not output change-related deep features, but they rather characterize each input independently. To cope these issues, some studies have recently proposed different architectures for supervised change detection in \ac{3D} \acp{PC} (see section \ref{sec:intro}). These architectures can therefore be used as a back-bone to our unsupervised method. In practice, both \textit{Siamese KPConv} \citep{degelis2023siamese}
    and \textit{Encoder Fusion SiamKPConv} \citep{degelis2023change}
    will be experimented. \textit{Siamese KPConv} architecture is chosen because it is the first architecture to perform change segmentation into raw \ac{3D} \acp{PC}. It extends the idea of Siamese networks with the KPConv convolution principle, as can be seen in Figure~\ref{fig:backbones}a. The very recent \textit{Encoder Fusion SiamKPConv} is experimented as well since it has been shown in \cite{degelis2023change} that it outperforms other state-of-the-art methods including \textit{Siamese KPConv}. The main idea behind this architecture is no more to encode the two \acp{PC} independently (as \textit{Siamese KPConv} do) but rather to include the change information along the encoding process, as can be seen in Figure~\ref{fig:backbones}b where the encoding in the bottom part takes into account also the differences between features.

    Thus, parameters $\theta$ to be optimized are parameters of these back-bone architectures.

\subsubsection{Use of a prototype layer}
In the original version of DeepCluster, the final classification layer of $g_{W}$ is re-initialized before each parameter optimization session (i.e., training steps) because there is no matching between two consecutive cluster assignments. A further improved version of DeepCluster (DeepCluster-V2) was proposed by \citet{caron2020unsupervised} where the classifier $g_W$ is replaced by the prototypes, i.e., cluster centers. This ends up with an explicit comparison of the features and the centroid matrix $C$, and  tends to improve stability and performance of DeepCluster. According to preliminary experiments, we also decided to use the centroid matrix $C$ defining pseudo-cluster centers. Therefore, the last fully connected layer parameters of the back-bone are set using the centroid matrix $C$. This so-called prototype layer is updated after each clustering step and fixed during the training step.

\subsubsection{Input data}
We aim at detecting changes into raw \ac{3D} \acp{PC}.
The back-bone (Siamese KPConv or Encoder Fusion SiamKPConv) is able to compute features directly in \ac{3D} \acp{PC}. However, given their large size, the \acp{PC} cannot be used directly to feed the network, and splitting the input data into smaller subsets is required (similarly to the decomposition of an image into patches in the 2D case). We follow here the strategy we already used in the supervised case \citep{degelis2023siamese,degelis2023change} and opt for a decomposition into (pairs of) vertical cylinders, that has been shown to be more relevant than (pairs of) spheres when \acp{PC} have a privileged orientation, as for airborne \ac{3D} \acp{PC} in our settings. Indeed, the changes are more prone to occur in the vertical direction (e.g., new building or demolition), as illustrated in Sec.~\ref{sec:res}. Please note that our methodology can be easily applied with other neighborhood shapes, such as spheres as already done for supervised cliff erosion monitoring \citep{degelis2022cliff}.

Furthermore, in their experimentation, \cite{caron2018deep} provides Sobel-filtered images as input to the \ac{CNN} instead of \ac{RGB} images. Sobel filtering acts as an edge detector thanks to the computation of gradients on the image. This seems an important step in their method \citep{caron2021self,mustapha2022deep} and acts as a pre-computation of relevant features. However, when dealing with \ac{3D} \acp{PC}, there is no direct equivalent to Sobel filtering.
Thus, we rely on some predefined features commonly used to characterize \ac{3D} \acp{PC}. Let us note that using such features can help the network, as already shown in a supervised scenario \citep{degelis2023change}. As such, we consider here some handcrafted features already used in a \ac{RF}-based change detection context \citep{tran2018integrated}:
    \begin{itemize}
    \item[\textbullet] Point distribution represented by point normals 
    and information on the distribution of points in the neighborhood (i.e., linearity, planarity, and omnivariance). 
    \item[\textbullet] Height information characterized by rank of the point on vertical axis in the neighborhood, maximum range of elevation of points in the neighborhood, and normalized height according to the local \ac{DTM} (rasterization of the \ac{PC} at the ground level). 
    \item[\textbullet] Change information described through a feature called stability (ratio of the number of points in the neighborhood of the current \ac{PC} to the number of points in the neighborhood in the other \ac{PC}).
\end{itemize}
We refer the reader to the original paper \citep{tran2018integrated} for a detailed description of these features.

    \subsubsection{Training considerations}
    Change segmentation task implies assigning a pseudo-label to each point of the second PC (of pairs of the training set). Considering the size of the training set, to fit in memory, a mini-batch $k$-means \citep{sculley2010web} clustering is used. The principle of splitting the largest cluster when an empty cluster appears is used as in DeepCluster.

    Change detection datasets are highly imbalanced. To avoid falling in a trivial solution where the back-bone predicts all points with the same label, after each clustering step weights $W_k$ (considering pseudo-labels distribution) are computed.
    These weights are further used to both select training cylinders and weight the loss ($\mathcal{L}_{NLL}$). Let us note that this cylinder selection process was also applied in the supervised context \citep{degelis2023siamese} (on the real labels though). It aims at giving more training samples of underrepresented pseudo-clusters. It also acts as a kind of data augmentation because from one epoch to another, selected cylinders differ according to the random drawing of the cylinder's central point. Without this trick, the method is likely to collapse to a single class prediction.

    During the training step, data augmentation appears to be crucial for stability and performance of the method. In particular, the following data augmentation strategies are used: random cylinders rotation around the vertical axis (same angle for both cylinders of a pair), and addition of a Gaussian noise at point level.

\subsection{From predicted pseudo-labels to real labels}\label{sec:DC3DCDmapping}
\begin{figure*}
    \centering
    \includegraphics[width=0.9\textwidth]{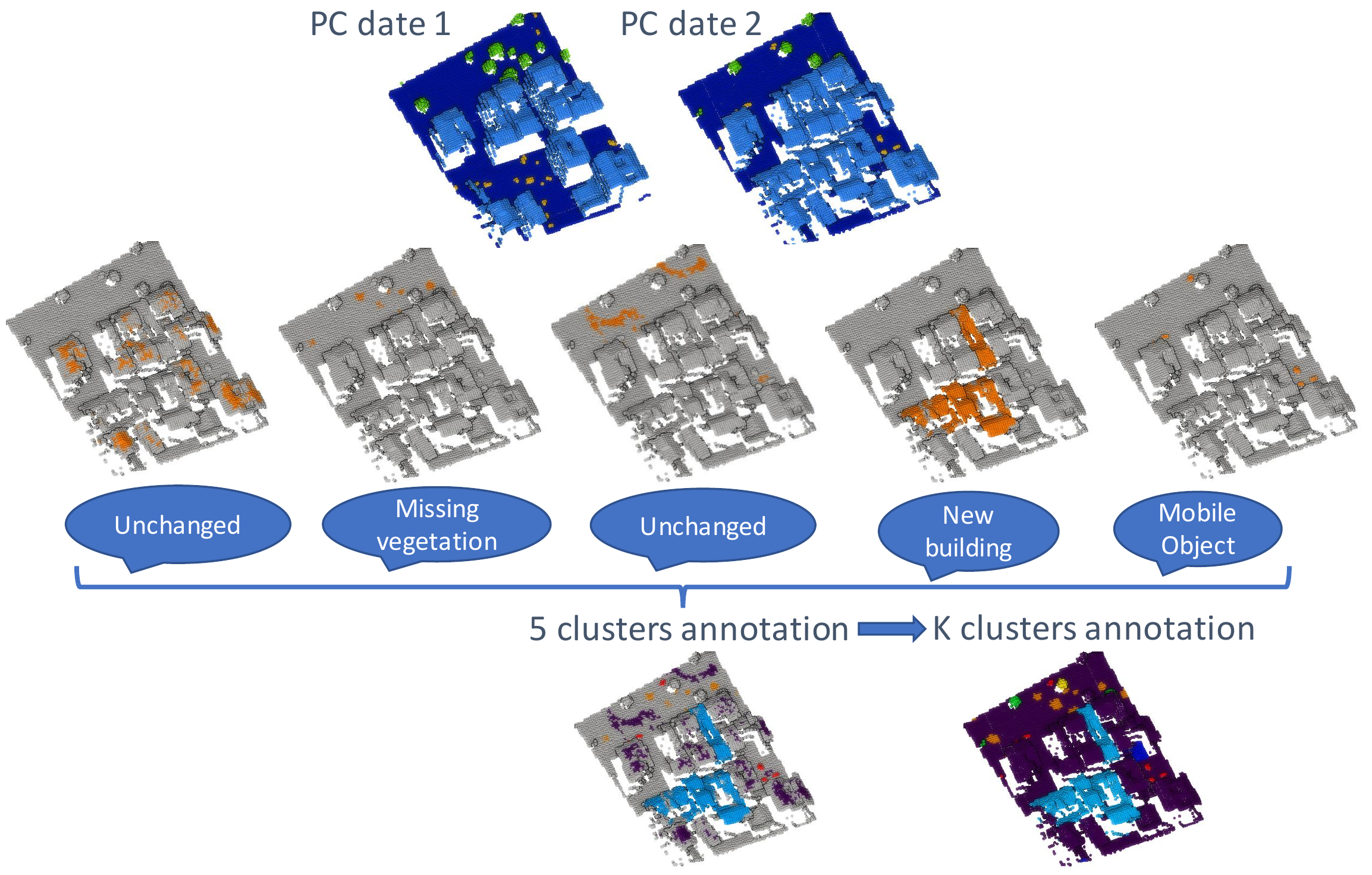}
    \caption[User guided mapping of predicted clusters to real classes]{\textbf{User guided mapping of predicted clusters to real classes.} For the $K$ predicted clusters, a mapping with the corresponding real class is performed by a user to obtain the final change segmentation of the PC. 5 mappings are provided for the sake of illustration. Segmenting the whole dataset requires $K$ annotations only. This is far less than the millions of points that need to be annotated in order to build training and validation sets in a supervised setting.}
    \label{fig:DC3DCDUserAnnot}
\end{figure*}

The above training using \ac{DC3DCD} method is fully unsupervised, thereby no use of a ground truth is required. At the end of the overall training process, the back-bone predicts labels for all points of the second PC according to the change. Predicted labels do not directly correspond to the real labels. There is an oversegmentation of PCs inducted by the choice of $K$, the number of pseudo-clusters, which is often large compared to the number of real classes.
By opting for such an oversegmentation setting, we expect to be able to address various use cases with different size and precision of classes. One real class is then composed of several predicted clusters, while we assume a predicted cluster to contain only one real class.
To map a real label onto each predicted label, a mapping step is necessary. For this mapping, we consider that the user should be involved in order to select the kind of changes that are of interest given the use case.  \ac{DC3DCD} enables to train a back-bone to segment the PC into small areas containing the same types of change or unchanged objects. Thus, the user just has to select for each predicted cluster a corresponding real class. It can be viewed as a kind of active learning process. This is illustrated in Figure~\ref{fig:DC3DCDUserAnnot}. This strategy finally involves $K$ annotations to obtain a final change segmentation over the whole testing set (no matter its size). $K$ corresponds to the number of pseudo-clusters used during the training. This hyper-parameter  has to be set beforehand. For this reason, our method can be viewed as weakly supervised since $K$ annotations by a user are required in the end of the process. However, these annotations are not taken into account in the learning process, and we therefore classify our approach in the family of unsupervised methods.

Note that although a labelling effort is required for this user-guided mapping, we argue that DC3DCD helps to greatly reduces this effort, as only $K$ annotations are necessary to predict changes throughout the complete dataset. In contrast, conventional deep network training requires annotations for millions of points.

In our experimental settings, we actually do not involve such a manual labelling step but mimic it through an automatic mapping strategy by relying on the ground truth labels provided with the public datasets. More precisely, each pseudo-cluster is assigned a real label by identifying its majority real class.
This method seems natural since in a really manual mapping the user typically examines the pseudo-cluster and assigns the most likely label, i.e., the class that corresponds to the majority of points within the pseudo-cluster.

\section{Experimental results}\label{sec:res}
In this part, we first present the experimental settings and protocol in Sec.~\ref{sec:expeprot}.
Since the proposed approach is unsupervised, a special attention must be given to the learning process, which behaves differently than in a more straightforward supervised situation. Therefore, we also analyze the learning behavior in Sec.~\ref{sec:DC3DCDPrelimRes}. We finally report our experimental results on both simulated and real data in Sec.~\ref{sec:DC3DCDUrb3DCD} and \ref{sec:DC3DCDresAHNCD}, respectively.

\subsection{Experimental settings and protocol}\label{sec:expeprot}
We detail here how we set the main hyper-parameters and we describe our experimental settings.

\subsubsection{Datasets}
Both simulated and real datasets will be experimented, considering Urb3DCD-V2 in low density LiDAR configuration \citep{degelis2021change, degelis2023siamese} and \ac{AHN-CD} \citep{degelis2023siamese}. To the best of our knowledge, these are the only public datasets for change segmentation. The \ac{Urb3DCD} dataset is composed of simulated \ac{ALS} \acp{PC} over a french city model (Lyon, France). A simulator is used in order to obtain an accurate annotation of \ac{3D} points concerning multiclass changes.
Conversely, \ac{AHN-CD} dataset is made of real \ac{ALS} \ac{PC} from \ac{AHN} campaigns over the Netherland \citep{sande2010assessment}.
Let us note that the registration of the \acp{PC} conducted by the Netherlands AHN services may contain some minor registration errors.
The change ground-truth is obtained thanks to a semi-automatic process. In \cite{degelis2023siamese}, it has been shown that this process is not perfect since it results in a lot of misclassifications. Therefore, the test set of this dataset will be used for qualitative assessment only. As for quantitative assessment, \ac{AHN-CD} comes with a manually annotated sub-part of the test set, on which we will report and compare quality scores.
Note that during the training, we make no use of the ground truth unless for the method assessment purpose (see Section~\ref{sec:DC3DCDPrelimRes}). The first sub-sampling rate $dl_0$ is set to \qty{1}{\m} and the cylinder radius to \qty{50}{\m} for Urb3DCD-V2-1. While for \ac{AHN-CD} $dl_0$ is set to \qty{0.5}{\m} and the cylinder radius to \qty{20}{\m} because of the difference of density between both datasets.

\subsubsection{Number of pseudo-clusters $K$}
As shown in different studies related to DeepCluster \citep{caron2018deep,mustapha2022deep}, choosing the number of pseudo-clusters is important.  We experimented several values for $K$  on the simulated dataset and found that $K=1000$ was an adequate compromise between the stability of the training process and the complexity of the segmentation (i.e,. the number of clusters).
Indeed, a too small value may not reflect all different types of changes (and thus does not allow the user to select the changes of interest), while a too large value leads to high training and annotation times.
The same value will also be used with the real dataset \ac{AHN-CD}.

\subsubsection{Training step and parameters optimization} For the training process, a \ac{SGD} with momentum of 0.98 is applied to minimize a point-wise NLL loss
using the pseudo-labels defined in the clustering step. A batch size of 10 is used. The initial learning rate is set to $10^{-3}$ and scheduled to decrease exponentially. As in \cite{caron2018deep}, we experimentally verified that reassigning the clustering after each epoch is better than an update after each $n$ epochs. Indeed, if several training epochs are conducted, the model seems to converge in the first local minimum associated with a non-optimal pseudo-clustering. In each epoch, \num{3000} cylinder pairs are seen by the model. A total of 55 epochs, i.e., 55 clustering and training steps, is performed.

\subsubsection{Comparisons with supervised methods}
A comparison with the only (to our knowledge) existing deep supervised methods is provided, namely \textit{Siamese KPConv} \citep{degelis2023siamese}, \textit{Encoder Fusion SiamKPConv} \citep{degelis2023change}, DSM-based deep learning methods (adaptation of \cite{daudt2018fully} networks to DSM inspired by \cite{zhang2019detecting}). Results of a \ac{RF} algorithm trained on hand-crafted features \citep{tran2018integrated} are also given.
For all these methods, the full training set available in the dataset (Urb3DCD-V2 or AHN-CD), made of millions of points, is used during the training step. Thus, these methods are expected to outperform our DC3DCD given the vast amount of additional labels they rely on. However, as already stated in the introduction, the literature is still very poor concerning unsupervised methods for change detection in 3D PCs. To the best of our knowledge, there is even no other unsupervised method for 3D PCs multi-class change detection, thus preventing us from conducting fair comparisons.

\subsubsection{Adaptation of a supervised method to a weakly supervised setting} To evaluate the benefit of our method, we propose a comparison  with deep learning-based supervised methods tuned to a weakly supervised setting. In practice, we use \textit{Encoder Fusion SiamKPConv}, the best of supervised techniques according to \cite{degelis2023change}. To this end, we trained it with the same amount of annotated data as our \ac{DC3DCD} setting. However, this is not straightforward since this network cannot be trained with only \num{1000} points.  Indeed, as during the supervised training of this network, labels should be provided for each  point of the second PC of the pair, and as a cylinder contains more than \num{1000} points (about \num{3500}), we cannot directly compare the supervised training with the same amount of labels (i.e., $K= 1000$). Thus, for both training and validation sets, we chose 7 cylinders, each one centered on one of the 7 classes contained in Urb3DCD-V2 to be sure that each class is represented. Note that with this minimal training configuration, the number of annotated points in the 14 cylinders is around \num{50000}. As 7 cylinders are less than the batch size of 10 used for all other deep learning-based methods, we also provide results with a batch size of 2.

\subsubsection{Comparisons with unsupervised methods} To the best of our knowledge, there is no other weakly supervised or unsupervised deep learning method tackling \ac{3D} \acp{PC} multiple change segmentation.
Thereby, we provide a comparison with a $k$-means algorithm applied on the ten hand-crafted features of \cite{tran2018integrated} dedicated to change detection in \ac{3D} \acp{PC}. Note that LiDAR specific features (e.g., intensity or number of echoes) used in \cite{tran2018integrated} are ignored here since the simulated dataset does not contain such information. For fair comparison, the $k$-means is set to predict also $K=1,000$ pseudo-clusters. The same user-guided mapping is done as proposed in the previous section (Figure~\ref{fig:DC3DCDUserAnnot}) to assign the final classes.

While there is no existing unsupervised deep learning method for multiclass change segmentation, the binary case was first and recently tackled in \citep{degelis2023deep}. In this setting, only two classes are considered (change and unchanged). In order to compare our method to other unsupervised deep learning models, we have decided to also including a comparison with the methods proposed in \citep{degelis2023deep}. They rely on \ac{DCVA} strategy to highlight binary change information. The deep features used in the comparison are extracted via trained models using a \ac{SSL} strategy in SSL-DCVA method, or a training on a semantic segmentation task using an available public dataset (namely \ac{H3D} \citep{kolle2021hessigheim}) in SSST-DCVA method. The binary change segmentation experiment is conducted only on the AHN-CD dataset, since the Urb3DCD dataset does not respect some hypothesis required for SSL-DCVA (we refer the reader to the original paper \citep{degelis2023deep} for more details).

Before presenting the quantitative results, we analyze in the next section the behavior of the network during the training process.

\subsection{Analysis of the learning process}\label{sec:DC3DCDPrelimRes}

Before presenting the results on the different datasets, we propose to study the behavior of \ac{DC3DCD} during the training phase. To do so, we rely on the \textit{Encoder Fusion SiamKPConv} back-bone and the configuration without the use of the ten hand-crafted features as input, so considering only 3D points coordinates. Note that the same tendencies are obtained with hand-crafted features or with \textit{Siamese KPConv} back-bone, but we prefer to show results with a network that takes into account the minimum information regarding changes. In practice, we compute criteria associated with the clustering quality and the pseudo-cluster distribution along the training process.

\subsubsection{Clustering quality}
The evolution of the clustering quality during training epochs is computed by comparing the pseudo-clusters obtained thanks to the $k$-means on deep features, and the real classes. More precisely, we compute the \ac{NMI} given by the following formula:
\begin{equation}\label{eq:nmi}
    \ac{NMI}(Y,Y_C) = \frac{I(Y,Y_C)}{\sqrt{H(Y)H(Y_C)}}
\end{equation}
where $Y$ and $Y_C$ contain the probabilities $p_i$, $p_{C_i}$, of each label $i=\{ 1 ... N\}$ associated with the true and pseudo-labels. $H$ is the entropy defined as:
\begin{equation}\label{eq:entropy}
    H(Y) = - \sum_{i=1}^{N}p_i \log_2 p_i
\end{equation}
and $I$ is the mutual information, defined as:
\begin{equation}
    I(Y,Y_C)= H(Y) - H(Y|Y_C)
\end{equation}
Intuitively, the \ac{NMI} is a measure of the information shared between two clusterings, i.e., in our case the clustering of deep features and real classes. If the \ac{NMI} is equal to 0, the two clusterings are totally independent. On the opposite, if the \ac{NMI} is equal to 1, there is a perfect correlation between the two clusterings, i.e., one of them is deterministically predictable from the other.

\begin{figure}
    \centering
    \begin{minipage}[t]{0.48\textwidth}
  \centering
  \centerline{\includegraphics[width=0.8\textwidth]{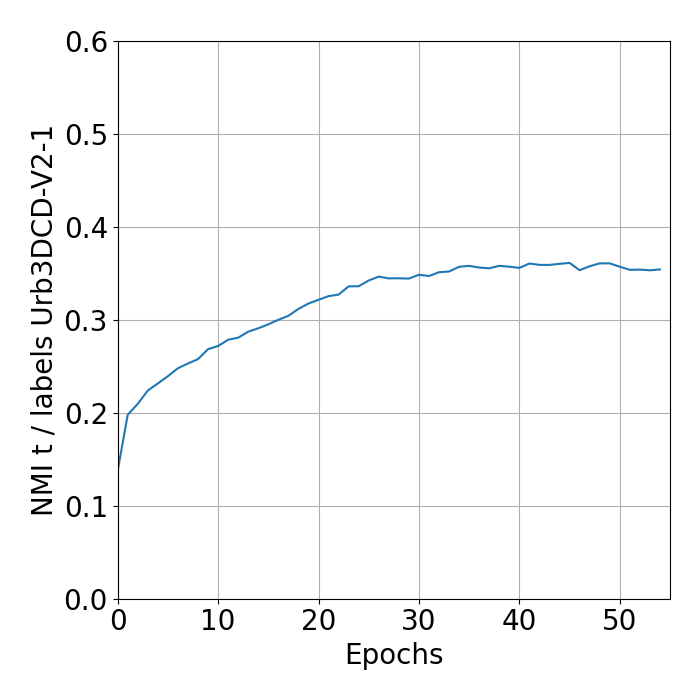}}
  \centerline{(\textbf{a}) Clustering quality }\medskip
    \end{minipage}
    \begin{minipage}[t]{0.48\textwidth}
  \centering
  \centerline{\includegraphics[width=0.8\textwidth]{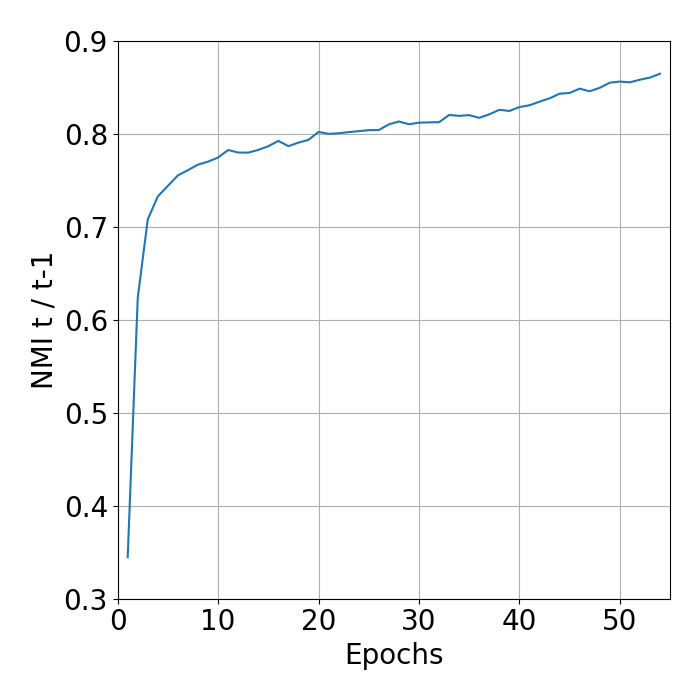}}
  \centerline{(\textbf{b}) Cluster reassignment }\medskip
    \end{minipage}
    \caption[Analysis of the behavior of \ac{DC3DCD} during the training.]{\textbf{Analysis of the behavior of \ac{DC3DCD} during the training.} The evolution of clustering quality (\textbf{a}) is given thanks to the \ac{NMI} between the clustering and the real labels of Urb3DCD-V2 dataset.  The \ac{NMI} between the clustering at epoch $t$ and the clustering at epoch $t-1$ gives the cluster reassignment (\textbf{b}).}
    \label{fig:DC3DCDnmi}
\end{figure}

We present the evolution of the clustering quality along the epochs in Figure~\ref{fig:DC3DCDnmi}a by giving the \ac{NMI} between the clustering and the real labels of Urb3DCD-V2 dataset. As can be seen, the clustering tends to get closer to real classes along with the training process. It seems to stabilize after 30-40 epochs. Let us remark that at the end of the training, the \ac{NMI} is around $0.35$. It is still far to 1, but the same trend was observed in DeepCluster training quality assessment by \cite{caron2018deep}. In Figure~\ref{fig:DC3DCDnmi}b, we evaluate the number of reassignments of cluster from one epoch to the following using the \ac{NMI} between the clustering of the two epochs. It seems that during the first epochs, there is an important evolution of the clustering, but the training  rapidly converges to a rather stable clustering ($NMI>0.8$). Again, the same tendency was obtained by \cite{caron2018deep} for DeepCluster on ImageNet dataset.

\subsubsection{Pseudo-cluster distribution}
\begin{figure}
    \centering
           \includegraphics[trim=0cm 0cm 0cm 0cm, clip, width=0.5\textwidth]{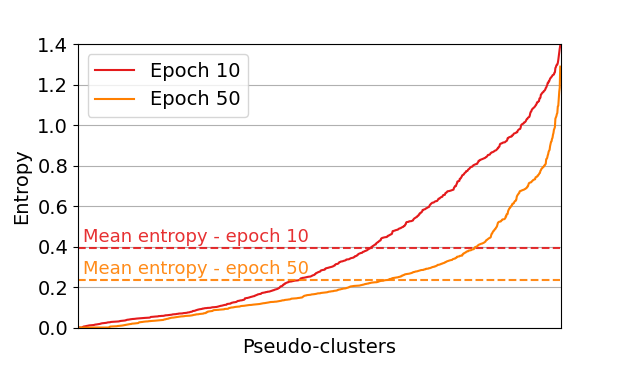}\\

    \caption{\textbf{Pseudo-cluster entropy at epoch 10 and 50.} The entropy is a measure of purity of pseudo-clusters. The lower the entropy, the purer the pseudo-cluster. Pseudo-clusters are sorted in increasing entropy values.}
    \label{fig:DC3DCDTrainrepart}
\end{figure}

As for the pseudo-cluster distribution, we remind that, ideally, a pseudo-cluster contains only one real class, and a real class can be distributed into several pseudo-clusters. To measure the purity of a pseudo-cluster, we investigate the entropy $H$ (Equation~\ref{eq:entropy}) of each pseudo-cluster. If it is near 0, then the pseudo-cluster contains almost only one real class. However, if a pseudo-cluster is divided into several classes, the entropy is higher. In Figure~\ref{fig:DC3DCDTrainrepart} is given the entropy for each pseudo-cluster at epoch 10 and 50. Pseudo-clusters are sorted in increasing entropy values. As can be seen, there is an improvement between epoch 10 and 50. The area under the entropy curves is indeed smaller at epoch 50 (0.24 of mean entropy) than at epoch 10 (0.39 of mean entropy), meaning that entropy values are globally smaller.
Following this first assessment, we have tried to understand better how the real classes were distributed within each pseudo-cluster. To do so, we focused on the majority class among each cluster, i.e., the most frequent label (among the real classes) in the set of \ac{3D} points belonging to the pseudo-cluster. We have observed that, after 50 epochs of training, for 80\% of the pseudo-clusters, the proportion of the majority class was higher than 93 \%.
These results confirmed the relevance of the proposed user-guided strategy to automatically map a pseudo-cluster onto the majority real class for the evaluation, as stated in the method description (see Section~\ref{sec:DC3DCDmapping}).

After having studied the training behavior of the \ac{DC3DCD} method, we will now compare it to the state-of-the-art on the testing set of both simulated and real datasets.

\subsection{Results on simulated Urb3DCD dataset}\label{sec:DC3DCDUrb3DCD}

Quantitative evaluation of \ac{DC3DCD} on the simulated Urb3DCD-V2-1 dataset is given in Table~\ref{tab:ResLid05DC3DCD} and Table~\ref{tab:DC3DCDResLid05Iou}. In these tables, we also recall supervised results for the sake of comparison. Let us first analyze \ac{DC3DCD} results without hand-crafted features (two first lines of the bottom part of Table~\ref{tab:ResLid05DC3DCD} and Table~\ref{tab:DC3DCDResLid05Iou}). As can be seen, results for both \textit{Siamese KPConv} and \textit{Encoder Fusion SiamKPConv} back-bones are rather low. Indeed, while requiring the same annotation effort, $k$-means algorithm trained on the same hand-crafted as the \ac{RF} method proposed in \cite{tran2018integrated} provides better results (cf. first line of middle part in Table~\ref{tab:ResLid05DC3DCD}). However, these results are interesting because the two experimented back-bones provide significantly different results. In particular, \textit{Encoder Fusion SiamKPConv} ends up with a mIoU$_{ch}$ \num{1.5} times higher than \textit{Siamese KPConv}. While in a supervised setting, the improvement of \textit{Encoder Fusion SiamKPConv} was of 5 points of mIoU$_{ch}$, in the unsupervised context the choice of the architecture seems even more crucial.

Then, when hand-crafted features are added to the input of the network, results are largely improved (cf. the two last lines of Table~\ref{tab:ResLid05DC3DCD}). While \ac{DC3DCD} with \textit{Siamese KPConv} architecture and hand-crafted features provides results comparable to the $k$-means algorithm, \ac{DC3DCD} yields compelling results with both hand-crafted features and the \textit{Encoder Fusion SiamKPConv} architecture. Indeed, in this configuration there is more than 15 points of mIoU$_{ch}$ of improvement compared to $k$-means. Furthermore, \ac{DC3DCD} with this configuration is better than a fully supervised RF, and provides results comparable with fully supervised deep architectures trained on 2.5D rasterization of \ac{3D} \acp{PC}. Thereby, providing hand-crafted features is an important step in unsupervised settings. One possible interpretation is that the unsupervised version is very tricky to train in reason of the large number of possible local minima. Adding hand-crafted features probably helps the initialization to be closer to the global minimum.

Notice that the \textit{Encoder Fusion SiamKPConv} in a weakly supervised setting provides rather low results given the higher annotation effort required (about 50,000 annotated points). Results with a batch size of 10 are not stable. This is explained by the fact that only one batch is seen per epoch, and the same learning rate scheduler as with a batch size of 2 is used. Thus, this training is more prone to fall in a local minimum. Even with a reduced batch size, leading to more stable results, we can see the benefit of using \ac{DC3DCD} for the training of  \textit{Encoder Fusion SiamKPConv} network because the effort of annotation is lower.

\begin{table*}
    \centering
    \footnotesize
    \begin{tabular}{cc|cc}
    \toprule
         &Method & mAcc (\%)& mIoU$_{ch}$  (\%)\\
        \hline
        \multirow{5}{*}{\rotatebox{90}{Supervised}}
        &Siamese KPConv  \citep{degelis2023siamese}& 91.21 $\pm$ 0.68 & 80.12 $\pm$ 0.02 \\
         &Encoder Fusion SiamKPConv \citep{degelis2023change} & \textbf{94.23} $\pm$ 0.88 & \textbf{85.19} $\pm$ 0.24\\

            &DSM-Siamese &  80.91 $\pm$ 5.29& 57.41 $\pm$ 3.77\\
            &DSM-FC-EF & 81.47 $\pm$ 0.55 &	56.98 $\pm$ 0.79 \\
            & \ac{RF} \citep{tran2018integrated} & 65.82 $\pm$ 0.05 & 52.37  $\pm$ 0.10 \\
           \hline
           \multirow{7}{*}{\rotatebox{90}{Weakly sup.}}

           &$k$-means  & 56.15 $\pm$ 0.62 & 41.46 $\pm$ 0.53\\
            &\textit{Encoder Fusion SiamKPConv} (batch size 10)
            & 29.03 $\pm$ 22.46 & 12.84 $\pm$ 18.49\\
            &\textit{Encoder Fusion SiamKPConv} (batch size 2) &53.09 $\pm$ 3.73 & 36.60 $\pm$ 3.18\\
            \cline{2-4}
           &\ac{DC3DCD} \textit{Siamese KPConv} &  28.28 $\pm$ 3.73 & 14.43 $\pm$ 3.70\\
            &\ac{DC3DCD} \textit{Encoder Fusion SiamKPConv} & 52.30 $\pm$ 2.41 & 37.75 $\pm$ 2.11\\
            &\ac{DC3DCD} \textit{Siamese KPConv} (with input features) & 54.91 $\pm$ 5.45 & 42.27 $\pm$ 6.64\\
            &\ac{DC3DCD} \textit{Encoder Fusion SiamKPConv} (with input features) & \textbf{68.45} $\pm$ 1.10 & \textbf{57.06} $\pm$ 0.41\\
        \bottomrule
    \end{tabular}\\
    \caption[Quantitative evaluation of \ac{DC3DCD} on Urb3DCD-V2 low density LiDAR dataset]{\textbf{Quantitative evaluation of \ac{DC3DCD} on Urb3DCD-V2 low density \ac{LiDAR} dataset.} \textit{Top}: supervised methods. DSM-based methods are adaptation of \cite{daudt2018fully} networks to \ac{DSM} inspired by \cite{zhang2019detecting} and \ac{RF} refers to Random Forests. \textit{Middle}: Weakly supervised methods with $k$-means and \textit{Encoder Fusion SiamKPConv} results using 7 training cylinders in the training and validation set (equivalent to about 50,000 annotated points). \textit{Bottom}: Weakly supervised methods with our proposed \ac{DC3DCD} evaluated in 4 different settings: with \textit{Siamese KPConv} or \textit{Encoder Fusion SiamKPConv} architectures and with or without the addition of 10 hand-crafted features as input to the network. }
    \label{tab:ResLid05DC3DCD}
\end{table*}

\begin{figure*}[ht]
    \centering
    \begin{tabular}{ccc}
        \includegraphics[width=0.27\textwidth]{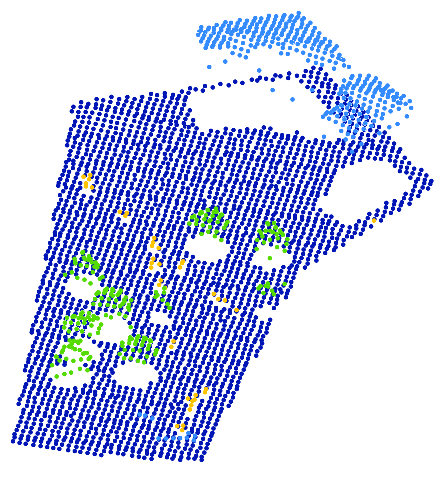} &
        \includegraphics[width=0.27\textwidth]{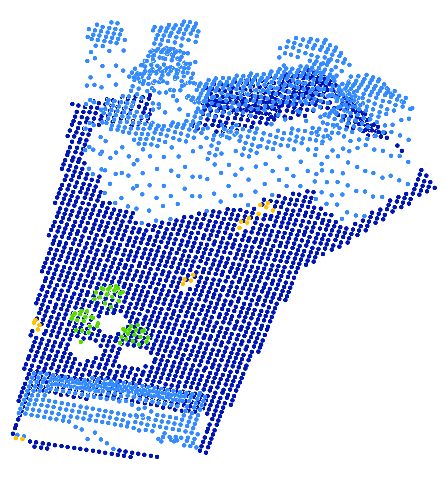} &
        \includegraphics[width=0.27\textwidth]{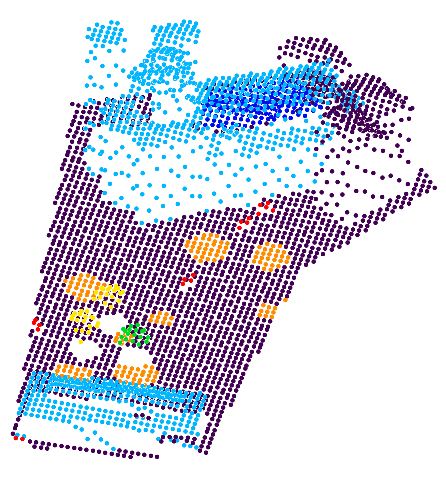}\\
        \multicolumn{2}{c}{        \begin{tikzpicture}
    		\begin{axis}[
            		xmin=1,
                    xmax=1.1,
                    ymin=1,
                    ymax=1.1,
                     hide axis,
    				width=0.5\textwidth ,
    				mark=circle,
    				scatter,
    				only marks,
    				legend cell align={left},
    				legend entries={Ground, Building, Vegetation, Mobile Objects},
    				legend style={draw=lightgray,at={(0,0)}, legend columns=4,/tikz/every even column/.append style={column sep=0.5cm}}]
    			\addplot[Ground] coordinates {(0,0)}; 
    			\addplot[Building] coordinates {(0,0)};
    			\addplot[Vegetation] coordinates {(0,0)};
    			\addplot[MO] coordinates {(0,0)};
    		\end{axis}
	    \end{tikzpicture}} & \\
        (\textbf{a}) PC 1 & (\textbf{b}) PC 2 & (\textbf{c}) GT \\

    \includegraphics[width=0.27\textwidth]{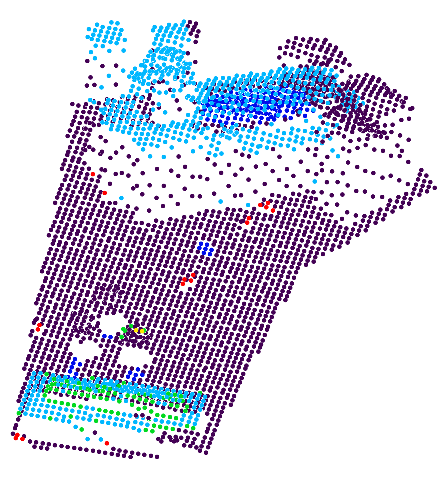}   &
        \includegraphics[width=0.27\textwidth]{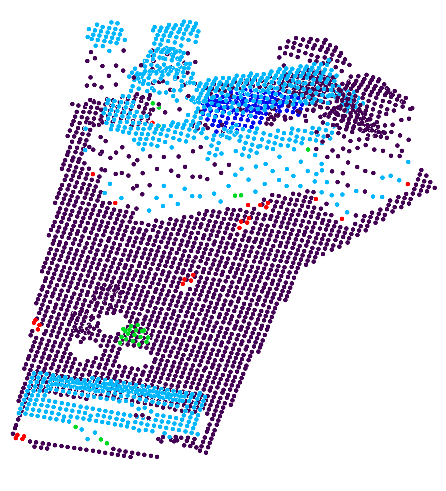}&
        \includegraphics[width=0.27\textwidth]{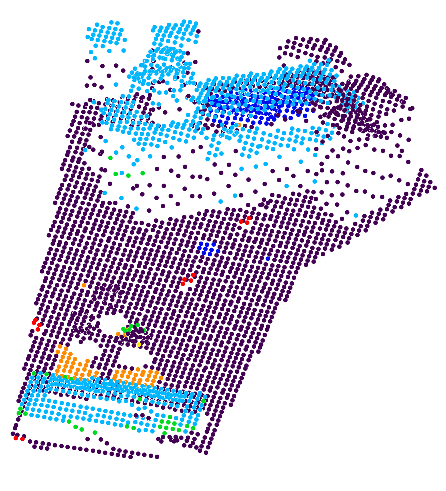}\\
        (\textbf{d}) $k$-means & (\textbf{e}) \ac{DC3DCD} EFSKPConv &(\textbf{f}) \ac{DC3DCD} EFSKPConv (input features)\\

        \multicolumn{3}{c}{        \begin{tikzpicture}
    		\begin{axis}[
            		xmin=1,
                    xmax=2,
                    ymin=1,
                    ymax=2,
                     hide axis,
    				width=0.5\textwidth ,
    				mark=circle,
    				scatter,
    				only marks,
    				legend entries={Unchanged, New Building, Demolition, New Vegetation, Vegetation Growth, Missing Vegetation, Mobile Objects},
    				legend cell align={left},
    				legend style={draw=lightgray,at={(0,0)}, legend columns=5,/tikz/every even column/.append style={column sep=0.2cm}}]
    			\addplot[Unchanged] coordinates {(0,0)}; 
    			\addplot[NewBuild] coordinates {(0,0)};
    			\addplot[Demol] coordinates {(0,0)};
    			\addplot[VegeN] coordinates {(0,0)};
    			\addplot[VegeG] coordinates {(0,0)};
    			\addplot[VegeR] coordinates {(0,0)};
    			\addplot[MOch] coordinates {(0,0)};
    		\end{axis}
	    \end{tikzpicture}}\\
    \end{tabular}

    \caption[Qualitative assessment of \ac{DC3DCD} method on Urb3DCD-V2 dataset (area~1)]{\textbf{Visual change detection results on Urb3DCD-V2 low density \ac{LiDAR} sub-dataset (area~1):} (\textbf{a}-\textbf{b}) the two input point clouds; (\textbf{c}) ground truth (GT): simulated changes; (\textbf{d}) $k$-means results; (\textbf{e}) \ac{DC3DCD} with the \textit{Encoder Fusion SiamKPConv} architecture results; (\textbf{f}) \ac{DC3DCD} with the \textit{Encoder Fusion SiamKPConv} architecture and the addition of 10 hand-crafted features as input results.}
    \label{fig:DC3DCDresLid05}
\end{figure*}
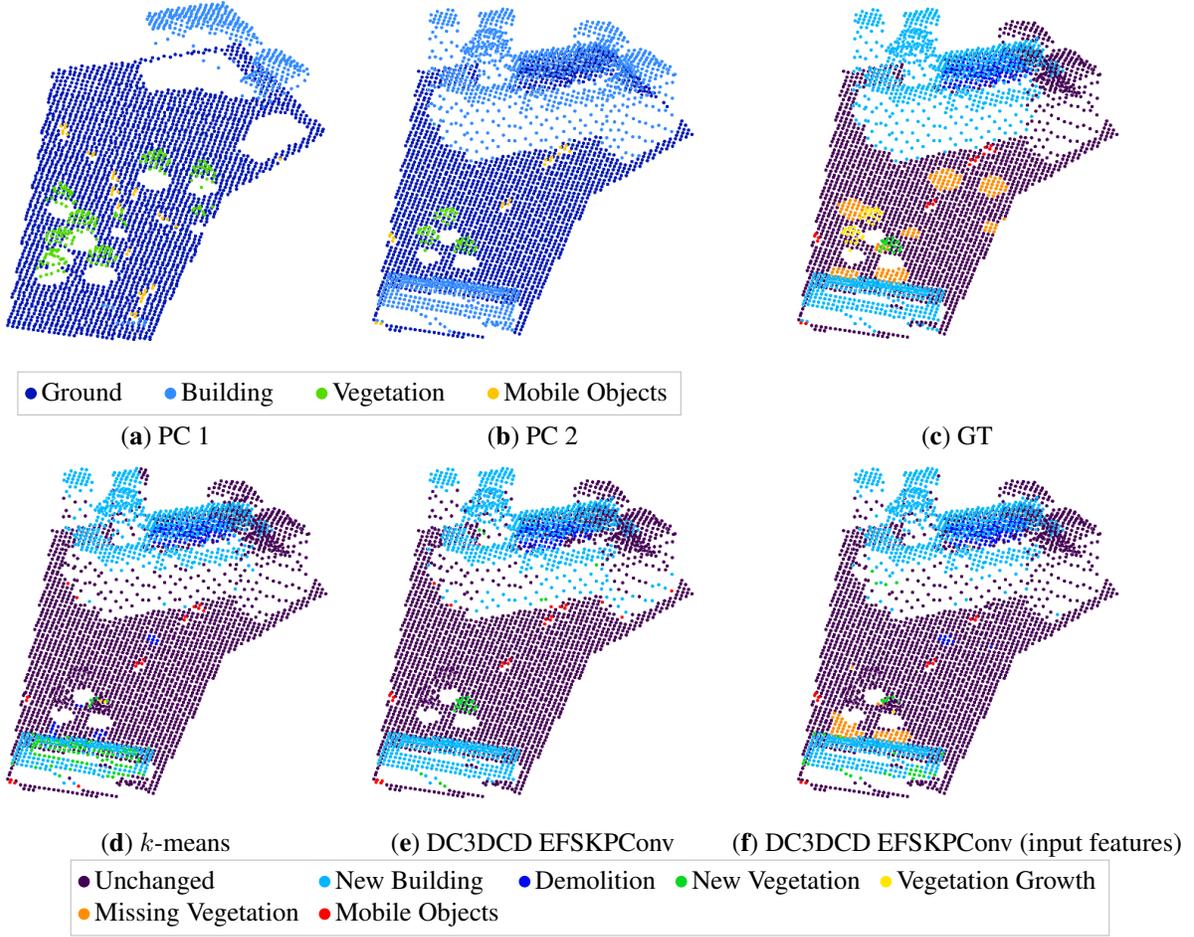

\begin{figure*}
    \centering
    \footnotesize
    \begin{tabular}{ccc}
        \includegraphics[width=0.27\textwidth]{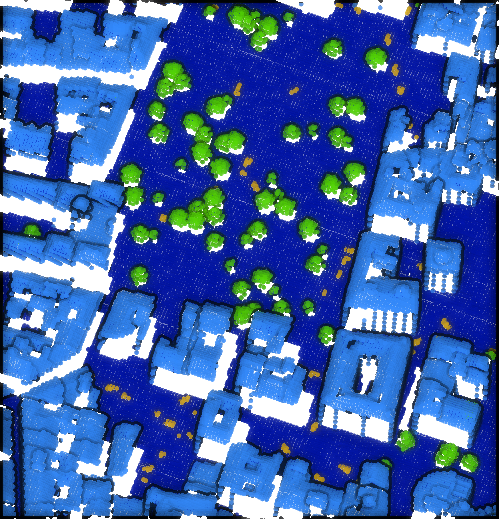} &
        \includegraphics[width=0.27\textwidth]{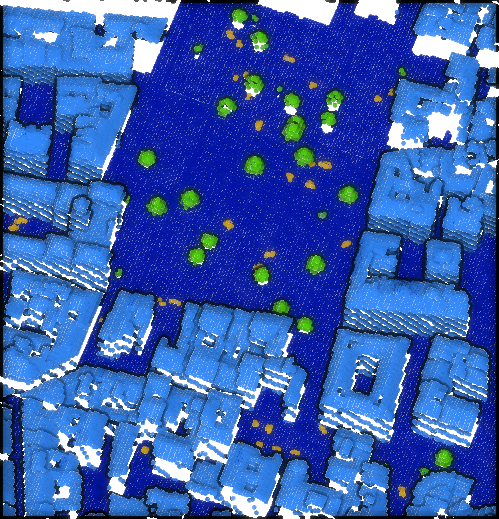} &
        \includegraphics[width=0.27\textwidth]{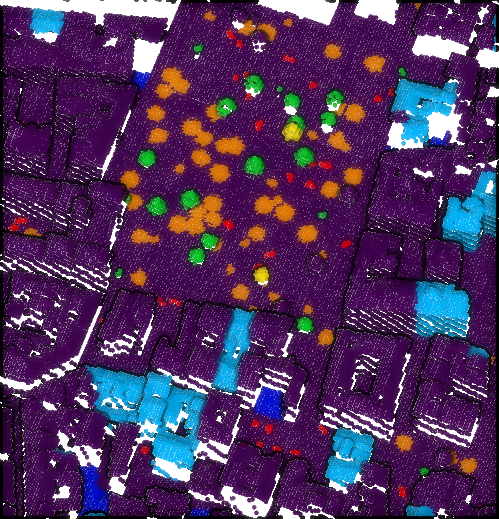}\\
        \multicolumn{2}{c}{        \begin{tikzpicture}
    		\begin{axis}[
            		xmin=1,
                    xmax=1.1,
                    ymin=1,
                    ymax=1.1,
                     hide axis,
    				width=0.5\textwidth ,
    				mark=circle,
    				scatter,
    				only marks,
    				legend cell align={left},
    				legend entries={Ground, Building, Vegetation, Mobile Objects},
    				legend style={draw=lightgray,at={(0,0)}, legend columns=5,/tikz/every even column/.append style={column sep=0.5cm}}]
    			\addplot[Ground] coordinates {(0,0)}; 
    			\addplot[Building] coordinates {(0,0)};
    			\addplot[Vegetation] coordinates {(0,0)};
    			\addplot[MO] coordinates {(0,0)};
    		\end{axis}
	    \end{tikzpicture}} & \\
        (\textbf{a}) PC 1 & (\textbf{b}) PC 2 & (\textbf{c}) GT \\

    \includegraphics[width=0.27\textwidth]{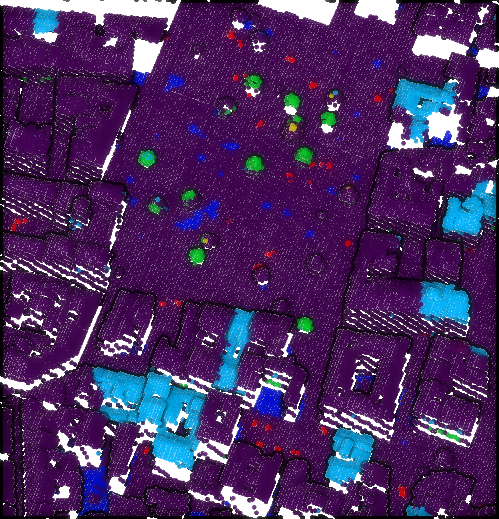}   &
        \includegraphics[width=0.27\textwidth]{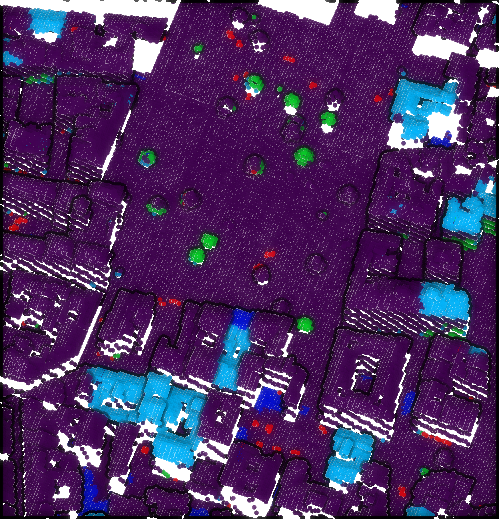}&
        \includegraphics[width=0.27\textwidth]{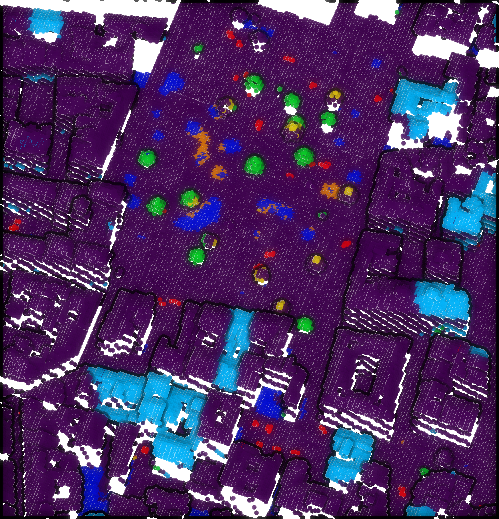}\\
        (\textbf{d}) $k$-means & (\textbf{e}) \ac{DC3DCD} EFSKPConv &(\textbf{f}) \ac{DC3DCD} EFSKPConv (input feat.)\\
        \multicolumn{3}{c}{        \begin{tikzpicture}
    		\begin{axis}[
            		xmin=1,
                    xmax=2,
                    ymin=1,
                    ymax=2,
                     hide axis,
    				width=0.5\textwidth ,
    				mark=circle,
    				scatter,
    				only marks,
    				legend entries={Unchanged, New Building, Demolition, New Vegetation, Vegetation Growth, Missing Vegetation, Mobile Objects},
    				legend cell align={left},
    				legend style={draw=lightgray,at={(0,0)}, legend columns=4,/tikz/every even column/.append style={column sep=0.2cm}}]
    			\addplot[Unchanged] coordinates {(0,0)}; 
    			\addplot[NewBuild] coordinates {(0,0)};
    			\addplot[Demol] coordinates {(0,0)};
    			\addplot[VegeN] coordinates {(0,0)};
    			\addplot[VegeG] coordinates {(0,0)};
    			\addplot[VegeR] coordinates {(0,0)};
    			\addplot[MOch] coordinates {(0,0)};
    		\end{axis}
	    \end{tikzpicture}}\\
    \end{tabular}

    \caption{\textbf{Visual change detection results on Urb3DCD-V2 low density \ac{LiDAR} sub-dataset (area~2):} (\textbf{a}-\textbf{b}) the two input point clouds; (\textbf{c}) ground truth (GT): simulated changes; (\textbf{d}) $k$-means results; (\textbf{e}) \ac{DC3DCD} with the \textit{Encoder Fusion SiamKPConv} architecture results; (\textbf{f}) \ac{DC3DCD} with the \textit{Encoder Fusion SiamKPConv} architecture and 10 hand-crafted input features.}
    \label{fig:DC3DCDresLid05-2}
\end{figure*}

Two different examples are given for a qualitative assessment of the method in Figures~\ref{fig:DC3DCDresLid05} and \ref{fig:DC3DCDresLid05-2}. As visible in Figure~\ref{fig:DC3DCDresLid05-2}, main changes (e.g., new buildings or demolitions) seem quite well retrieved by the $k$-means and both \ac{DC3DCD} configurations. However, when going more into details, some misclassifications can be seen on new building facades (Figure~\ref{fig:DC3DCDresLid05}) or vegetation. For new building facades, a slight improvement over $k$-means is reached by DC3DCD, but it is still not perfect. The $k$-means technique has the same tendency as the \ac{RF} method (see \cite{degelis2023siamese}) and confuses small new buildings with new vegetation, surely because they have the same height as visible in Figure~\ref{fig:DC3DCDresLid05}. As depicted in Table~\ref{tab:DC3DCDResLid05Iou}, main difficulties of the \ac{DC3DCD} method concern vegetation growth and missing vegetation. Note that this was already the most difficult classes in the supervised context. The missing vegetation is almost always confused with demolition in \ac{DC3DCD} with hand-crafted input features and the \textit{Encoder Fusion SiamKPConv} architecture. This is even worse with the $k$-means and missing vegetation is never predicted with \ac{DC3DCD} without hand-crafted input features. However, this make sense, since the
`missing vegetation' and `demolition' classes are both negative changes. Surprisingly, mobile objects are quite well retrieved, especially for \ac{DC3DCD} with hand-crafted input features and the \textit{Encoder Fusion SiamKPConv} architecture (Table~\ref{tab:DC3DCDResLid05Iou}).
\begin{table*}
    \centering
    \scriptsize
        \begin{tabular}{cl|ccccccc}
        \toprule
        & & \multicolumn{7}{c}{Per class IoU (\%)}\\
        & Method & Unchanged & New building & Demolition & New veg. & Veg. growth& Missing veg.& Mobile Object\\
        \hline
        \multirow{5}{*}{\rotatebox{90}{Supervised}}
        &SKPConv \tiny{\citep{degelis2023siamese}} & 95.82 $\pm$ 0.48 & 86.68 $\pm$ 0.47 & 78.66 $\pm$ 0.47 & 93.16 $\pm$ 0.27 & 65.17  $\pm$ 1.37 & 65.46  $\pm$ 0.93 & 91.55 $\pm$ 0.60\\
           &EFSKPConv \tiny{\citep{degelis2023change}} & \textbf{97.47} $\pm$ 0.04 & \textbf{96.68} $\pm$ 0.30 & \textbf{82.29} $\pm$ 0.16 & \textbf{96.52} $\pm$ 0.03 & \textbf{67.76} $\pm$ 1.51 & \textbf{73.50} $\pm$ 0.81 &\textbf{ 94.37} $\pm$ 0.54\\
           & DSM-Siamese &  93.21 $\pm$ 0.11& 86.14 $\pm$ 0.65 & 69.85 $\pm$ 1.46& 70.69 $\pm$ 1.35& 8.92 $\pm$ 15.46 &	60.71 $\pm$ 0.74& 8.14 $\pm$ 5.42\\
           & DSM-FC-EF & 94.39  $\pm$ 0.12 &	91.23  $\pm$ 0.31 & 71.15 $\pm$ 0.99 & 68.56 $\pm$ 3.92 & 1.89 $\pm$ 2.82 & 62.34 $\pm$ 1.23 & 46.70 $\pm$ 3.49 \\
            & \ac{RF} \tiny{\citep{tran2018integrated}}& 92.72 $\pm$ 0.01 & 73.16 $\pm$ 0.02 & 64.60 $\pm$ 0.06 & 75.17 $\pm$ 0.06 & 19.78 $\pm$ 0.30 & 7.78 $\pm$ 0.02 & 73.71  $\pm$ 0.63 \\
        \hline
        \multirow{7}{*}{\rotatebox{90}{Weakly sup.}}

           &$k$-means & 91.82 $\pm$ 0.05 & 70.46 $\pm$ 0.25 & 59.83 $\pm$ 0.11 & 59.20 $\pm$ 0.48 & 6.00 $\pm$ 0.19 & 0.00 $\pm$ 0.00 & 53.26 $\pm$ 3.19\\
            & EFSKPConv (b. s. 10)
            & 58.38 $\pm$ 46.55 & 13.22 $\pm$ 16.83 & 26.15 $\pm$ 23.54& 11.04 $\pm$ 19.12 & 0.41 $\pm$ 0.71 & 7.29 $\pm$ 11.99 & 19.48 $\pm$ 33.46\\
             & EFSKPConv (b. s. 2) & 89.88 $\pm$ 0.53 & 29.26 $\pm$ 16.83 &  48.66 $\pm$ 2.57 & 52.91 $\pm$ 9.19 & 7.59 $\pm$ 6.40 & 14.35 $\pm$ 15.16 & 66.84 $\pm$ 5.42 \\
             \cline{2-9}
           &\ac{DC3DCD} SKPConv &  84.51 $\pm$ 0.70 & 13.33 $\pm$ 3.05 & 29.50 $\pm$ 13.19 & 40.31 $\pm$ 9.15 & 3.03 $\pm$ 1.80 & 0.08 $\pm$ 0.01 & 0.33 $\pm$ 0.29\\
            &\ac{DC3DCD} EFSKPConv &90.90 $\pm$ 0.79 & 64.06 $\pm$ 5.13 & 54.35 $\pm$ 3.84 & 58.14 $\pm$ 20.03 & 1.45 $\pm$ 2.05 & 0.94 $\pm$ 0.78 & 47.57 $\pm$ 2.58\\
            &\ac{DC3DCD} SKPConv (i. f.) & 92.90 $\pm$ 0.21 & 76.61 $\pm$ 2.09 & 67.22 $\pm$ 2.63 & 61.33 $\pm$ 10.07 & 8.66 $\pm$ 6.54 & 16.39 $\pm$ 14.95 &23.44 $\pm$ 40.54\\
            &\ac{DC3DCD} EFSKPConv (i. f.) & \textbf{93.96} $\pm$ 0.11 & \textbf{79.26} $\pm$ 0.68 & \textbf{67.88} $\pm$ 0.49 & \textbf{75.34} $\pm$ 2.81 & \textbf{19.48} $\pm$ 4.00 &\textbf{ 20.29} $\pm$ 2.90 & \textbf{80.10} $\pm$ 3.16\\
             \bottomrule
        \end{tabular}
    \caption[Per-class \ac{IoU} scores of \ac{DC3DCD} on Urb3DCD-V2 low density LiDAR dataset]{\textbf{Per-class \ac{IoU} scores of \ac{DC3DCD} on Urb3DCD-V2 low density \ac{LiDAR} dataset.} \textit{Top}: supervised methods. DSM-based methods are adaptation of \cite{daudt2018fully} networks to \ac{DSM} inspired by \cite{zhang2019detecting} and \ac{RF} refers to Random Forests. \textit{Middle}: Weakly supervised methods with $k$-means and \textit{Encoder Fusion SiamKPConv} results using 7 training cylinders in the training and validation set (equivalent to about 50,000 annotated points). \textit{Bottom}: Weakly supervised methods with our proposed \ac{DC3DCD} evaluated in 4 different settings: with \textit{Siamese KPConv} (SKPConv) or \textit{Encoder Fusion SiamKPConv} (EFSKPConv) architectures and with or without the addition of 10 hand-crafted features as input to the network.  Veg. stands for vegetation; b. s. for batch size; i.f. for input features.}
    \label{tab:DC3DCDResLid05Iou}
\end{table*}

\subsection{Results on real \ac{AHN-CD} dataset}\label{sec:DC3DCDresAHNCD}

\ac{Urb3DCD} being a synthetic dataset, additional experiments are needed to assess the behavior of our method on real data. We consider here the \ac{AHN-CD} dataset in two settings. First, we conduct a multi-class change segmentation scenario as done previously with \ac{Urb3DCD}. Then, to allow comparison with other unsupervised deep models for \ac{3D} \ac{PC} change detection, we reduce the multi-class scenario to a bi-class problem, where the results consist only in the presence or absence of a change (and not the type of change).

\subsubsection{Multi-class change segmentation}
\begin{figure*}[ht]
    \centering
    \begin{tabular}{ccc}
        \includegraphics[width=0.3\textwidth]{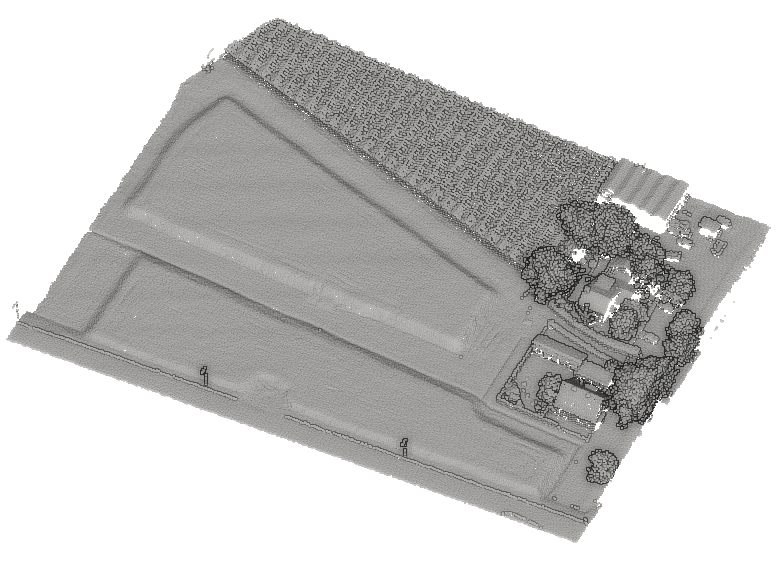} &
        \includegraphics[width=0.3\textwidth]{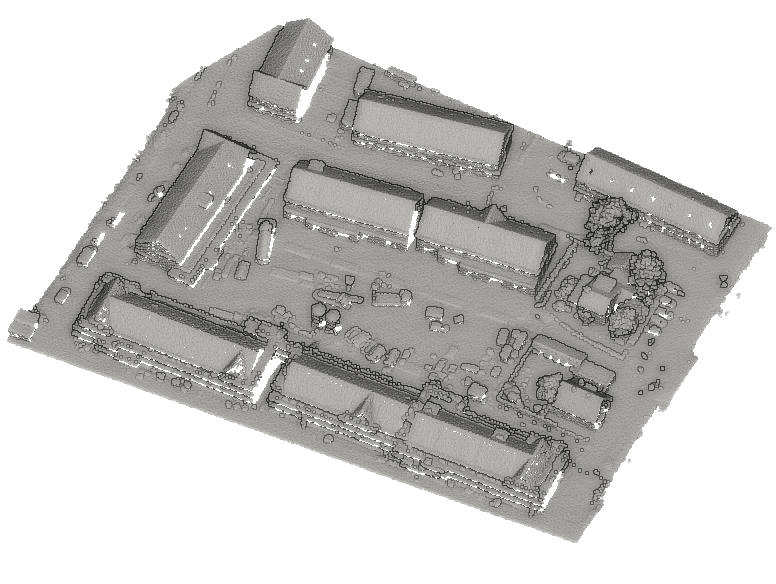} &
        \includegraphics[width=0.3\textwidth]{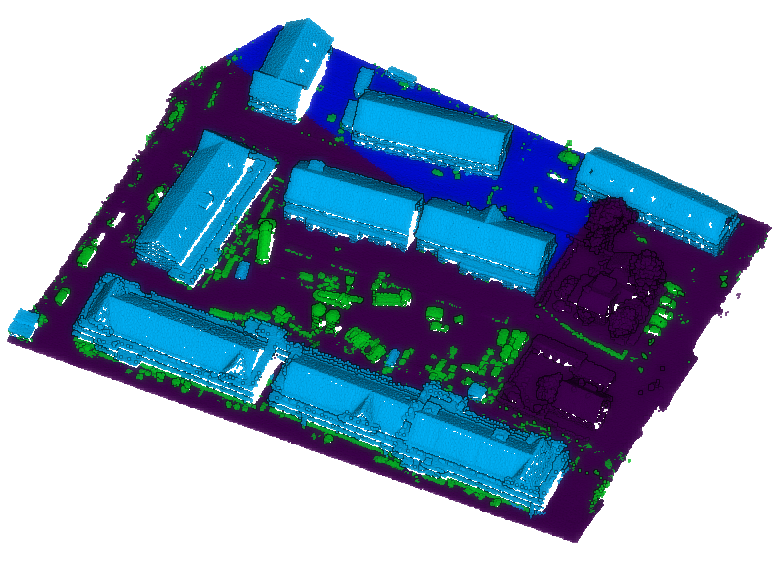} \\
         (\textbf{a}) PC date 1 & (\textbf{b}) PC date 2 & (\textbf{c}) Ground truth \\
    \end{tabular}
        \begin{tabular}{cc}
           \includegraphics[width=0.3\textwidth]{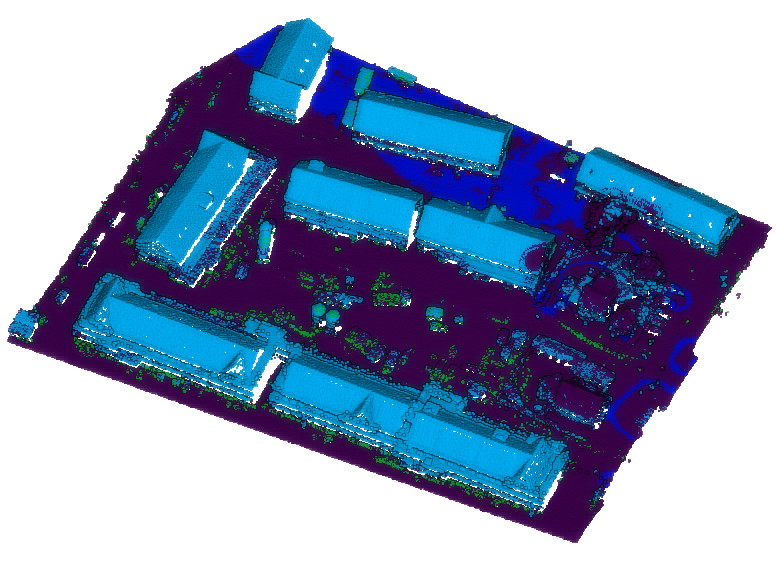} &
        \includegraphics[width=0.3\textwidth]{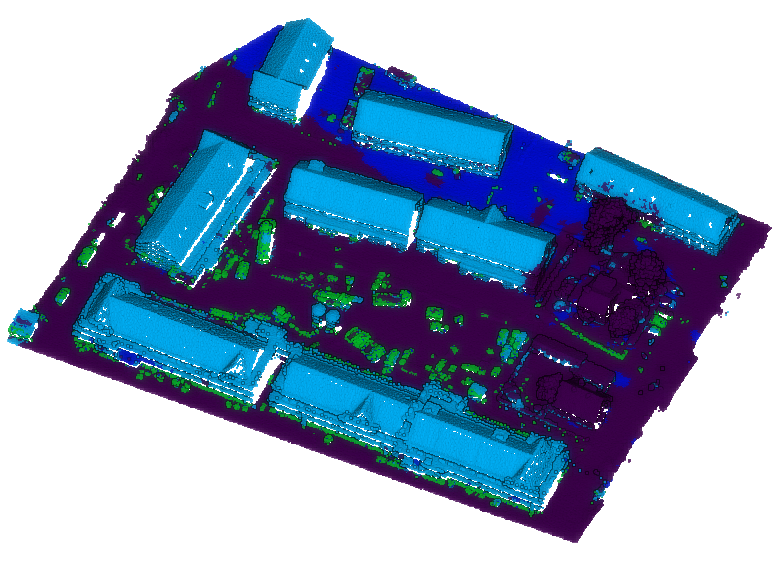} \\
          (\textbf{d}) $k$-means & (\textbf{e}) \ac{DC3DCD} EFSKPConv (i. f.) \\
                  \multicolumn{2}{c}{
        \begin{tikzpicture}
    		\begin{axis}[
            		xmin=1,
                    xmax=2,
                    ymin=1,
                    ymax=2,
                     hide axis,
    				width=0.5\textwidth ,
    				mark=circle,
    				scatter,
    				only marks,
    				legend entries={Unchanged, New Building, Demolition, New Clutter},
    				legend cell align={left},
    				legend style={draw=lightgray,at={(0,0)}, legend columns=4,/tikz/every even column/.append style={column sep=0.5cm}}]
    			\addplot[Unchanged] coordinates {(0,0)}; 
    			\addplot[NewBuild] coordinates {(0,0)};
    			\addplot[Demol] coordinates {(0,0)};
    			\addplot[VegeN] coordinates {(0,0)};
    		\end{axis}
	    \end{tikzpicture}}\\
	   \begin{tikzpicture}
    \node (image) at (0,0) { \includegraphics[width=0.3\textwidth]{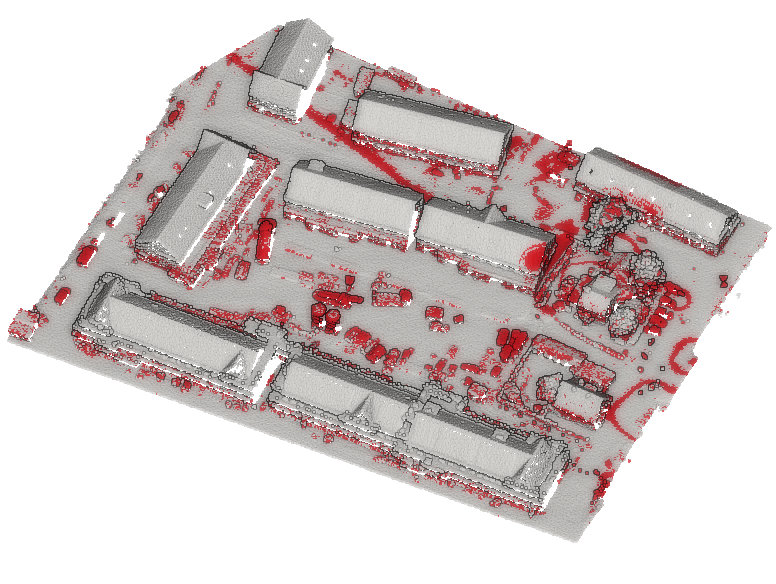}};
       \draw[black, very thick,rotate=50] (1.1,-0.8) ellipse (16pt and 10pt);
      \draw[black, very thick,rotate=75] (-0.4,-1.4) ellipse (6pt and 8pt);
 \end{tikzpicture}
	    &
        \includegraphics[width=0.3\textwidth]{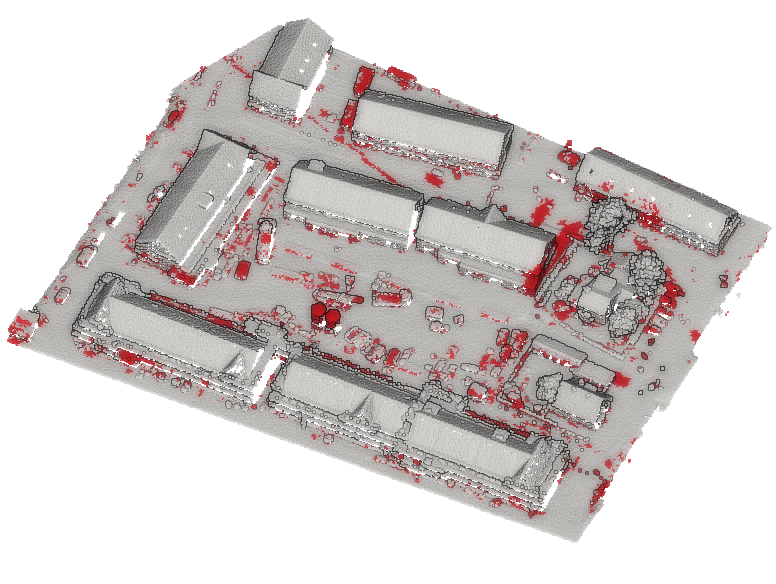} \\
          (\textbf{f}) $k$-means errors & (\textbf{g}) \ac{DC3DCD} EFSKPConv (i. f.) errors \\
          \multicolumn{2}{c}{ \begin{tikzpicture}
    		\begin{axis}[
            		xmin=1,
                    xmax=2,
                    ymin=1,
                    ymax=2,
                     hide axis,
    				width=0.5\textwidth ,
    				mark=circle,
    				scatter,
    				only marks,
    				legend entries={GT differences},
    				legend cell align={left},
    				legend style={draw=lightgray,at={(0,0)}, legend columns=1,/tikz/every even column/.append style={column sep=0.5cm}}]
    			\addplot[red] coordinates {(0,0)}; 
    		\end{axis}
	    \end{tikzpicture} }
    \end{tabular}
    \caption[Qualitative results on the manually annotated sub-part of \ac{AHN-CD} dataset]{\textbf{Qualitative results on the manually annotated sub-part of \ac{AHN-CD} dataset:} (\textbf{a}-\textbf{b}) PCs at date 1 and 2; (\textbf{c}) ground truth;  $k$-means results (\textbf{d}) and errors (\textbf{f}); \ac{DC3DCD} results (\textbf{e}) and errors (\textbf{g}) using the \textit{Encoder Fusion SiamKPConv} architecture and the 10 hand-crafted features in input. Regions of interest specifically discussed in the text are highlighted with ellipses. }
    \label{fig:DC3DCDAHNCDres_cm}
\end{figure*}
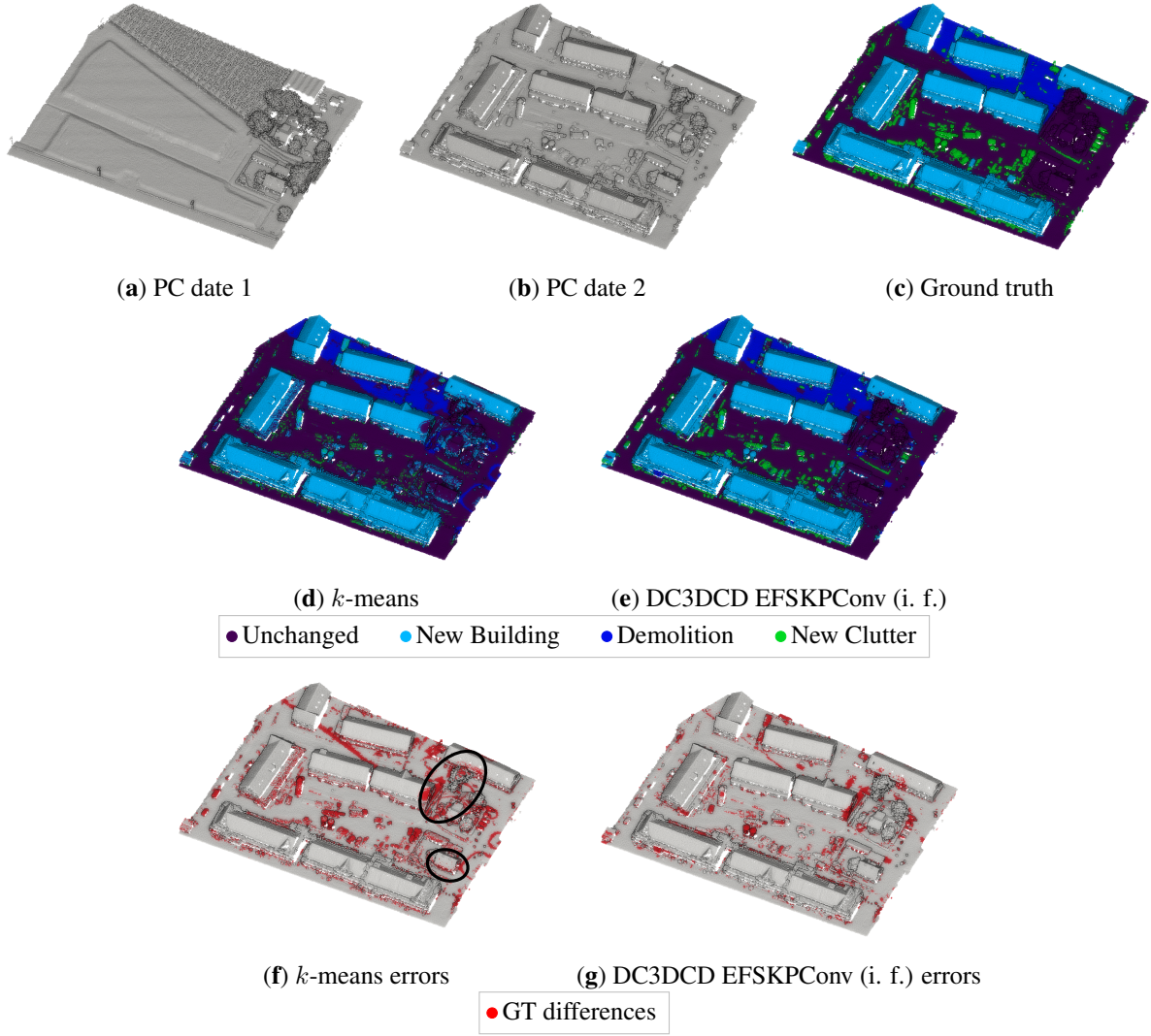

\begin{figure*}[ht]
\footnotesize
    \centering
    \begin{tabular}{ccc}
        \includegraphics[width=0.3\textwidth]{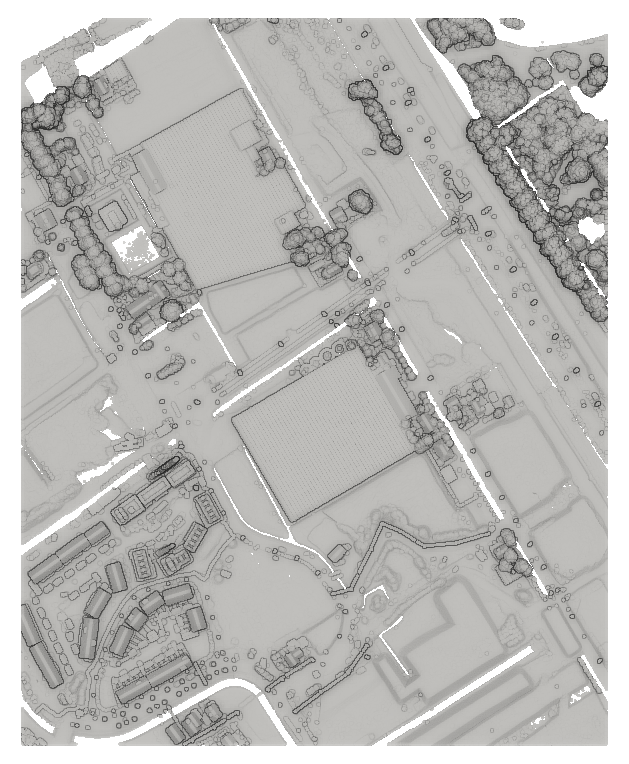} &
        \includegraphics[width=0.3\textwidth]{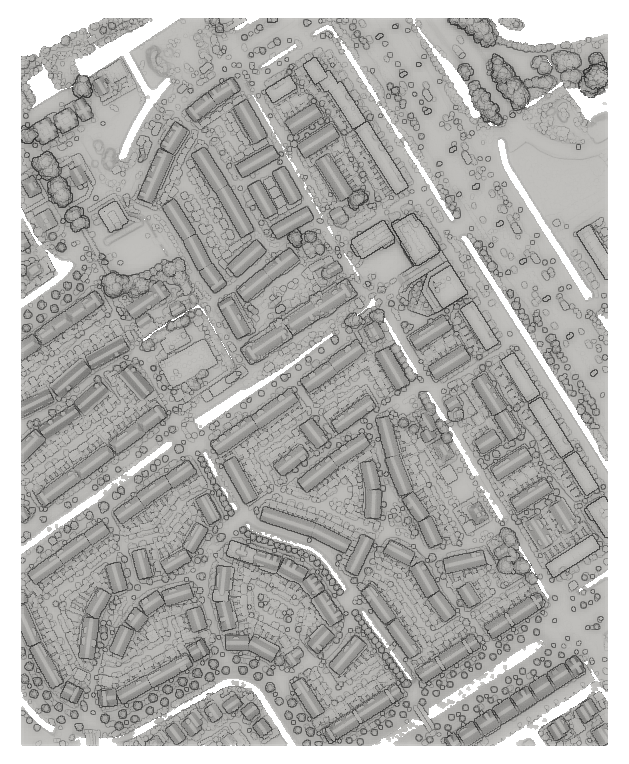} &
        \includegraphics[width=0.3\textwidth]{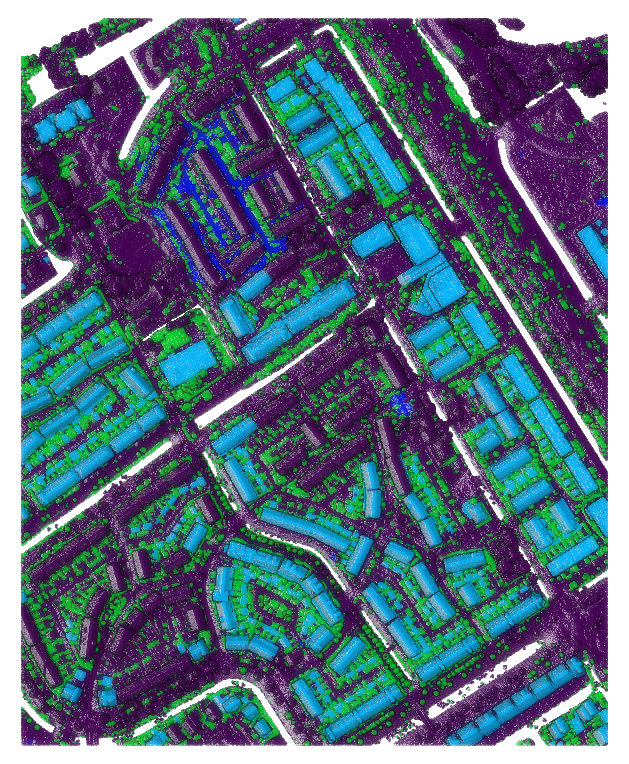} \\
         (\textbf{a}) PC date 1 & (\textbf{b}) PC date 2 & (\textbf{c}) Ground truth \\
\begin{tikzpicture}
    		\begin{axis}[
            		xmin=1,
                    xmax=2,
                    ymin=1,
                    ymax=2,
                     hide axis,
    				width=0.5\textwidth ,
    				mark=circle,
    				scatter,
    				only marks,
    				legend entries={Unchanged, New Building, Demolition, New Clutter},
    				legend cell align={left},
    				legend style={draw=lightgray,at={(0,0)}, legend columns=1,/tikz/every even column/.append style={column sep=0.5cm}}]
    			\addplot[Unchanged] coordinates {(0,0)}; 
    			\addplot[NewBuild] coordinates {(0,0)};
    			\addplot[Demol] coordinates {(0,0)};
    			\addplot[VegeN] coordinates {(0,0)};
    		\end{axis}
	    \end{tikzpicture}&
        \includegraphics[width=0.3\textwidth]{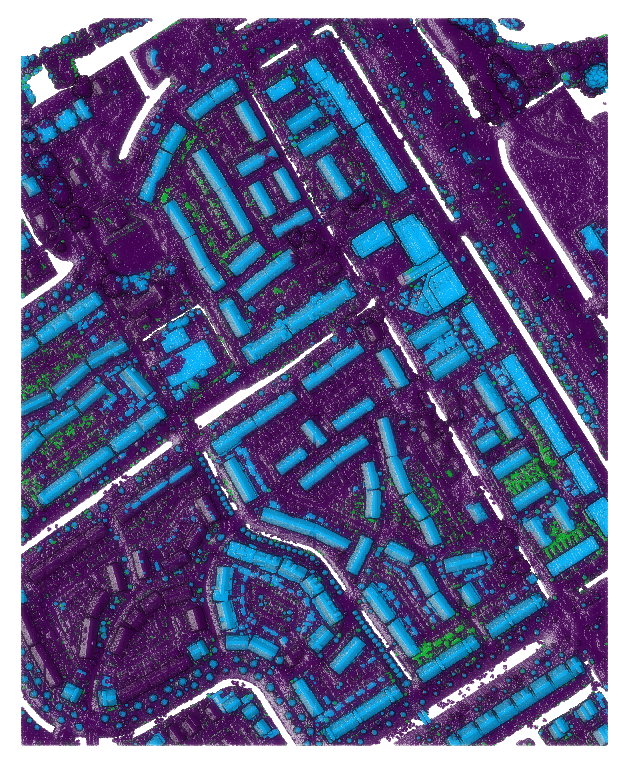} &
        \includegraphics[width=0.3\textwidth]{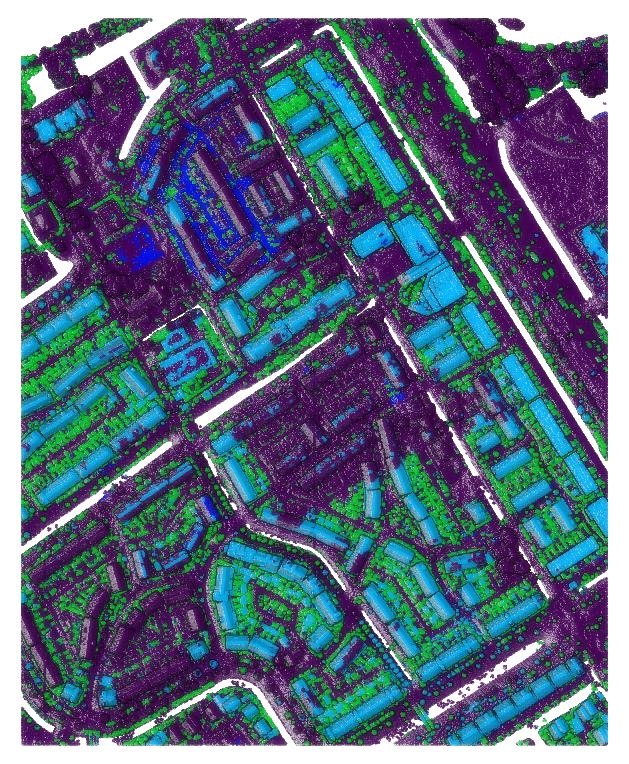} \\
          &(\textbf{d}) $k$-means & (\textbf{e}) \ac{DC3DCD} EFSKPConv (i. f.) \\

	        \begin{tikzpicture}
    		\begin{axis}[
            		xmin=1,
                    xmax=2,
                    ymin=1,
                    ymax=2,
                     hide axis,
    				width=0.5\textwidth ,
    				mark=circle,
    				scatter,
    				only marks,
    				legend entries={GT differences},
    				legend cell align={left},
    				legend style={draw=lightgray,at={(0,0)}, legend columns=1,/tikz/every even column/.append style={column sep=0.5cm}}]
    			\addplot[red] coordinates {(0,0)}; 
    		\end{axis}
	    \end{tikzpicture}    & \begin{tikzpicture}
    \node (image) at (0,0) {\includegraphics[width=0.3\textwidth]{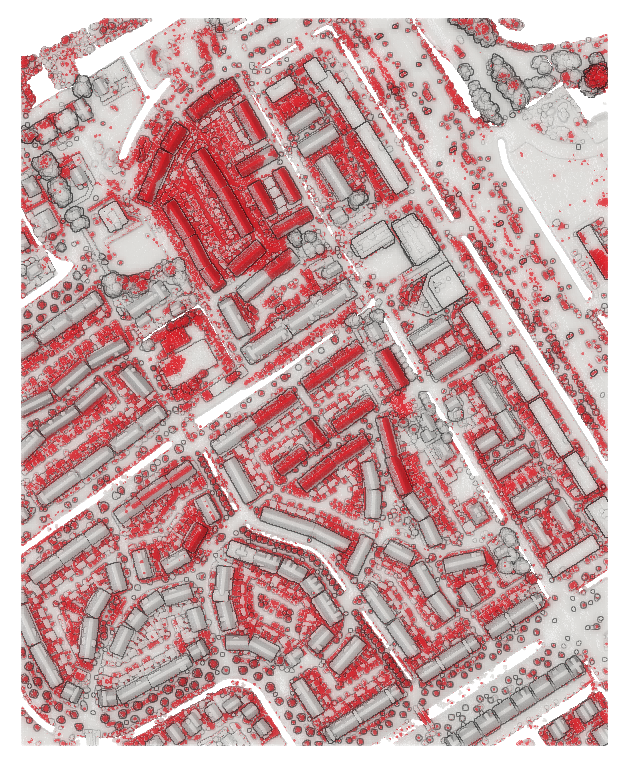}};
    \draw[black,very thick] (-1.5,2.3) rectangle (0.2,0.7);
     \draw[black, very thick] (-0.8,0.4) rectangle (0.9,-1);
 \end{tikzpicture} &
        \begin{tikzpicture}
    \node (image) at (0,0) {\includegraphics[width=0.3\textwidth]{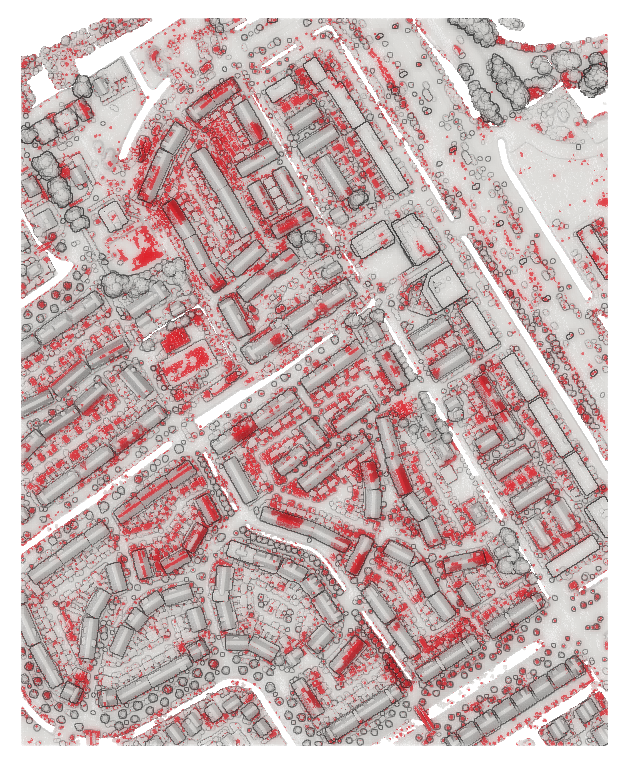}};
    \draw[black, very thick] (-1.5,2.3) rectangle (0.2,0.7);
    \draw[black, very thick] (-0.8,0.4) rectangle (0.9,-1);
 \end{tikzpicture} \\
          &(\textbf{f}) $k$-means errors & \makecell{(\textbf{g}) \ac{DC3DCD} EFSKPConv (i. f.)\\ errors}\\
    \end{tabular}
    \caption[Qualitative results on the semi-automatically annotated \ac{AHN-CD} dataset]{\textbf{Qualitative results on the semi-automatically annotated \ac{AHN-CD} dataset:} (\textbf{a}-\textbf{b}) PCs at date 1 and 2; (\textbf{c}) ground truth;  $k$-means results (\textbf{d}) and corresponding errors  (\textbf{f});  \ac{DC3DCD} results (\textbf{e}) and corresponding errors  (\textbf{g}) using the \textit{Encoder Fusion SiamKPConv} architecture and the 10 hand-crafted features in input. Regions of interest specifically discussed in the text are highlighted with rectangles. }
    \label{fig:DC3DCDAHNCDres_full}
\end{figure*}

Concerning the real \ac{AHN-CD} dataset, quantitative results  on the manually annotated testing set are given in Table~\ref{tab:DC3DCDresAHN_manualClean}, and Table~\ref{tab:DC3DCDresAHN_clean_iou} for per class results.
While this dataset may contain some registrations errors, we have not observed any significant effect on the change segmentation results.
For comparison purpose, we also provide results of supervised methods. However, we recall that they have been trained on the semi-automatically annotated \ac{AHN-CD} dataset containing several ground truth errors \citep{degelis2023siamese}. This explains lower results of the \ac{RF} compared to the $k$-means which have been mapped onto real classes using the manually annotated set (as for \ac{DC3DCD} method). As already observed with the simulated dataset, we can see that \ac{DC3DCD} provides better results than the $k$-means algorithm. Figure~\ref{fig:DC3DCDAHNCDres_cm} shows that main changes are well retrieved for both  methods. However, in the $k$-means results, larger objects of the clutter class such as trucks are mixed up with buildings. There are also lots of misclassifications in unchanged vegetation and unchanged building facades (see region of interest in Figure~\ref{fig:DC3DCDAHNCDres_cm}f). As far as \ac{DC3DCD} is concerned, unchanged vegetation is well classified. A few mistakes are visible in some
`new clutter' objects. We recall that this class is a mix of a lot of objects, from vegetation to cars or garden sheds, surely explaining why its classification score is lower. Complementary results on a larger \ac{AHN-CD} testing tile (\qtyproduct{1 x 1.2}{\km} corresponding to ca. 65 million points) are shown in Figure~\ref{fig:DC3DCDAHNCDres_full}. The ground truth is given by the semi-automatic process detailed in \cite{degelis2023siamese}. The mapping onto the real classes is performed using this ground truth for both $k$-means method and DC3DCD. As visible in this example, most of
`new clutter' class objects are omitted or mixed up with the new building class, also the demolition class is totally omitted by the $k$-means algorithm (Figure~\ref{fig:DC3DCDAHNCDres_full}d). In the \ac{DC3DCD} results in Figure~\ref{fig:DC3DCDAHNCDres_full}e, clutter class seems better retrieved, even though it is not perfect implying the main differences with the ground truth (Figure~\ref{fig:DC3DCDAHNCDres_full}g). In the areas of interest depicted by the black rectangles in Figure~\ref{fig:DC3DCDAHNCDres_full}, we observe that \ac{DC3DCD} seems to better adapt to the user context (i.e., by the ground truth defined by the user) than the $k$-means, even though the same ground truth-guided mapping step has been performed. Indeed, here buildings are not set as new in the ground truth and in DC3DCD, conversely to $k$-means. Finally, on this tile, if we compare to the ground truth, \ac{DC3DCD} obtains 55.91\% of mIoU$_{ch}$, while the $k$-means only 24.63\%.
\begin{table*}
    \centering
    \footnotesize
    \begin{tabular}{cl|cc}
    \toprule
          &Method & mAcc (\%) &mIoU$_{ch}$ (\%) \\
        \hline
        \multirow{5}{*}{\rotatebox{90}{Supervised}}
            &Siamese KPConv \citep{degelis2023siamese}& 85.65 $\pm$ 1.55 & 72.95 $\pm$ 2.05 \\
            &Encoder Fusion SiamKPConv \citep{degelis2023change}& \textbf{90.26} $\pm$ 0.22 &\textbf{ 75.00} $\pm$ 0.74 \\
            & DSM-Siamese& 50.87 $\pm$ 1.15 & 30.96 $\pm$ 2.48 \\
            &DSM-FC-EF & 71.47 $\pm$ 1.43 & 45.57 $\pm$ 0.98 \\
            & \ac{RF} \citep{tran2018integrated} & 47.94 $\pm$ 0.02 & 29.45 $\pm$ 0.02 \\
            \hline
        \multirow{2}{*}{\rotatebox{90}{WS}}
            & $k$-means & 70.07 $\pm$ 0.56  & 53.12 $\pm$ 0.79\\
           & \ac{DC3DCD} \textit{Encoder Fusion SiamKPConv} (with input features) &\textbf{ 83.18} $\pm$ 1.10 &\textbf{ 66.69} $\pm$ 2.19\\

    \bottomrule
    \end{tabular}
    \caption[Qualitative assessment of \ac{DC3DCD} on the manually annotated sub-part of \ac{AHN-CD} dataset ]{\textbf{Qualitative assessment of \ac{DC3DCD} on the manually annotated sub-part of \ac{AHN-CD} dataset.} \textit{Top}: supervised methods.  DSM-based methods are adaptation of \cite{daudt2018fully} networks to \ac{DSM} inspired by \cite{zhang2019detecting} and \ac{RF} refers to Random Forests. In supervised settings, the training is performed on the semi-automatically annotated \ac{AHN-CD} dataset containing some errors (see \cite{degelis2023siamese}). \textit{Bottom}: Weakly supervised methods with  $k$-means and our proposed \ac{DC3DCD} with \textit{Encoder Fusion SiamKPConv} architecture and with the addition of 10 hand-crafted features as input to the network.}
    \label{tab:DC3DCDresAHN_manualClean}
\end{table*}
\begin{table*}
    \centering
    \scriptsize
    \begin{tabular}{cl|cccc}
    \toprule
          &\multirow{2}{*}{Method} & \multicolumn{4}{c}{Per class IoU (\%)}\\
          & &  Unchanged & New building & Demolition & New clutter\\
        \hline
                \multirow{5}{*}{\rotatebox{90}{Supervised}}
            & Siamese KPConv \citep{degelis2023siamese} &  89.75 $\pm$ 2.18 & 82.77 $\pm$ 5.38 & 86.44 $\pm$ 0.88 & \textbf{46.65} $\pm$ 0.16\\
            & Encoder Fusion SiamKPConv \citep{degelis2023change}& \textbf{94.79} $\pm$ 0.34 & \textbf{95.31} $\pm$ 1.95 & \textbf{88.87} $\pm$ 1.59 & 41.16 $\pm$ 1.30\\
            &DSM-Siamese&  77.10 $\pm$ 1.51 & 76.77 $\pm$ 0.79 & 4.91 $\pm$ 8.33 & 11.20 $\pm$ 1.71\\
            &DSM-FC-EF &70.77 $\pm$ 1.13 & 90.32 $\pm$ 0.61 & 30.58 $\pm$ 1.76 & 15.81 $\pm$ 0.81\\
            & \ac{RF} \citep{tran2018integrated}& 78.24 $\pm$ 0.00 & 74.64 $\pm$ 0.03 & 0.00 $\pm$ 0.00 & 13.72 $\pm$ 0.06\\
             \hline
             \multirow{2}{*}{\rotatebox{90}{WS}}
             &$k$-means & 84.13 $\pm$ 0.49 & 83.13 $\pm$ 0.89 & 55.40 $\pm$ 0.50 & 20.84 $\pm$ 1.00\\
             & \ac{DC3DCD} \textit{Encoder Fusion SiamKPConv} (with input features) &\textbf{91.34} $\pm$ 1.21 & \textbf{89.91} $\pm$ 0.72 & \textbf{69.52} $\pm$ 4.97 & \textbf{40.63} $\pm$ 0.97\\
             \bottomrule
    \end{tabular}
    \caption[Per class IoU \ac{DC3DCD} results on \ac{AHN-CD} dataset.]{\textbf{Per class \ac{IoU} \ac{DC3DCD} results on the manually annotated testing part of \ac{AHN-CD} dataset} given in \%. \textit{Top}: supervised methods. In supervised settings, the training is performed on the semi-automatically annotated \ac{AHN-CD} dataset containing some errors (see \cite{degelis2023siamese}). \textit{Bottom}: Weakly supervised methods with  $k$-means and our proposed \ac{DC3DCD} with \textit{Encoder Fusion SiamKPConv} architecture and with the addition of 10 hand-crafted features as input to the network.}
    \label{tab:DC3DCDresAHN_clean_iou}
\end{table*}

\subsubsection{Binary change detection}

To enable comparison with other state-of-the-art methods in unsupervised deep learning, we conduct some experiments in a binary change segmentation (a.k.a. change detection) setup. The results for the manually annotated sub-part of  \ac{AHN-CD} are given in Table~\ref{tab:compaUnsupAllBinary}. For these results, \ac{DC3DCD}  employs user-guided mapping directly for the binary setup (changed and unchanged classes). As evident from these quantitative results, \ac{DC3DCD}  outperforms other fully unsupervised methods significantly. However, it does require \num{1000} annotations for mapping the pseudo-clusters to the real classes. Our weakly supervised approach appears as a relevant compromise between supervised methods (here Siamese KPConv and Encoder Fusion SiamKPConv) providing high accuracy at the cost of very expensive annotation (millions of points) and unsupervised ones (here SSL-DCVA and SSST-DCVA) that do not need any annotations but achieve much lower accuracy.

\begin{table*}[]
    \centering
    \footnotesize
    \begin{tabular}{cl|c|cc}
 \toprule
&\multirow{2}{*}{Method}  &mAcc &  \multicolumn{2}{c}{IoU (\%)}\\
&  & (\%) & Unchanged & Changed \\
\midrule
 \multirow{2}{*}{\rotatebox{90}{Sup.}}
  & Siamese KPConv \citep{degelis2023siamese} & \textbf{97.08} & \textbf{95.39} & \textbf{92.95} \\
  & Encoder Fusion SiamKPConv \citep{degelis2023change}& 96.75  &  94.79 & 92.10 \\
\midrule
 \multirow{2}{*}{\rotatebox{30}{Unsup.}}
 & SSL-DCVA \citep{degelis2023deep}& \textbf{85.20} & \textbf{78.91} & \textbf{69.38}\\
& SSST-DCVA \citep{degelis2023deep}& 81.88 & 70.02 &63.85\\
\midrule
W. Sup & DC3DCD \textit{Encoder Fusion SiamKPConv} (with input features) & \textbf{94.43} & \textbf{91.24} &\textbf{86.96}\\
\bottomrule
\end{tabular}
    \caption{\textbf{Quantitative comparison on a binary change segmentation task on the manually annotated sub-part of AHN-CD dataset.}}
    \label{tab:compaUnsupAllBinary}
\end{table*}


\section{Discussion}\label{sec:discu}
In this paper, we have proposed an unsupervised change detection method with a weakly user-guided mapping to real classes providing convincing results. Unsupervised 3D PCs change detection is still open and complex and in the following, we point out some observations and discussions about possible improvements.

\subsection{Importance of network's architectures and input features}
We saw in the result section that the choice of the back-bone architecture and the addition of hand-crafted features as input along with 3D point coordinates are crucial. This is in agreement with the original publication of DeepCluster, where image gradients are provided as input to obtain accurate results \citep{caron2018deep}. These results in an unsupervised context also emphasizes conclusions of \cite{degelis2023change} on the necessity of applying convolution on features difference. To explain this, let us note that the unsupervised context is a largely unconstrained problem. While the annotation allows counterbalancing architectures weaknesses, this is indeed no longer possible for the unsupervised setting. Thereby the choice of an architecture that more specially extracts change-related features through convolutions of difference of features from both inputs at multiple scales, and the addition of well-designed hand-crafted features, allows guiding the training of the network toward a relevant minimum, leading to an accurate change segmentation.

\subsection{Improving \ac{DC3DCD} with contrastive learning}
As mentioned, the problem in an unsupervised setting is to train a network to extract appropriate  features for a specific task. In a task involving comparison of similar and dissimilar data (like change detection task), the contrastive loss is often used to force the network to extract identical features for similar data. Therefore, we explore here how to force the network to predict similar features for unchanged areas. To do so,
we propose to introduce the following contrastive term in the loss function:
\begin{equation}\label{eq:contloss}
\begin{array}{ccc}
    \mathcal{L}_{cont} = 0.5 \times y_{sim} \times F_{CD}^{2} & \mbox{ with }  & y_{sim} = \left\{
                \begin{array}{ll}
                    1 & \mbox{if similar}  \\
                    0 & \mbox{else.}
                \end{array}
            \right.
    \end{array}
\end{equation}
where $F_{CD}$ is the $L_2$-norm of output features and $y_{sim}$ is the similarity term (set to 1 for unchanged points, and 0 elsewhere). Using this contrastive term in the loss aims at forcing to 0 change-related features in unchanged areas. To test this idea, we combine the contrastive loss in Equation~\ref{eq:contloss} with the deep clustering loss (\ac{NLL} using the pseudo-label) taking the mean, and train the \textit{Encoder Fusion SiamKPConv} network since it gave the best results.

We first carried out experiments by taking the similarity $y_{sim}$ from the ground truth (as $y_{sim}$ is not available in practice, we first test the idea by taking real values of $y_{sim}$). Results were really convincing since, as visible in Table~\ref{tab:ResLid05DC3DCDV2Appendix} and Table~\ref{tab:DC3DCDResLid05IouV2Appendix}, \ac{DC3DCD} reached $73.51\%$ of mIoU$_{ch}$ without the use of hand-crafted features and $82.63\%$ of mIoU$_{ch}$  with the use of hand-crafted features on Urb3DCD-V2-1 dataset. We recall that on this dataset and in a fully supervised setting, \textit{Siamese KPConv} and \textit{Encoder Fusion SiamKPConv} networks obtained 80.12\% and 85.19\% of mIoU$_{ch}$ respectively. Thus, the addition of the contrastive part allows meeting fully supervised results (in an ideal case where $y_{sim}$ is known).

This first experience validated the idea of using the contrastive loss. However in practice, $y_{sim}$ needs to be estimated. To obtain the similarity $y_{sim}$, we first used the significant changes given by \ac{M3C2} \citep{lague2013accurate} or a binary thresholding of \ac{C2C} distance \citep{girardeau2005change}. Results are mitigated (see Table~\ref{tab:ResLid05DC3DCDV2Appendix} and Table~\ref{tab:DC3DCDResLid05IouV2Appendix}) since, without hand-crafted input features, the best results obtained using \ac{M3C2} for $y_{sim}$ allow us to improve by only 2\% the mean of \ac{IoU} over classes of change obtained with \ac{DC3DCD}. With hand-crafted input features, results are worsened when the contrastive term is added during the training (using $y_{sim}$ based on \ac{M3C2}, obtained mIoU$_{ch}$ is 46.42\%).

\begin{figure*}[ht]
    \centering
    \includegraphics[width=0.8\textwidth]{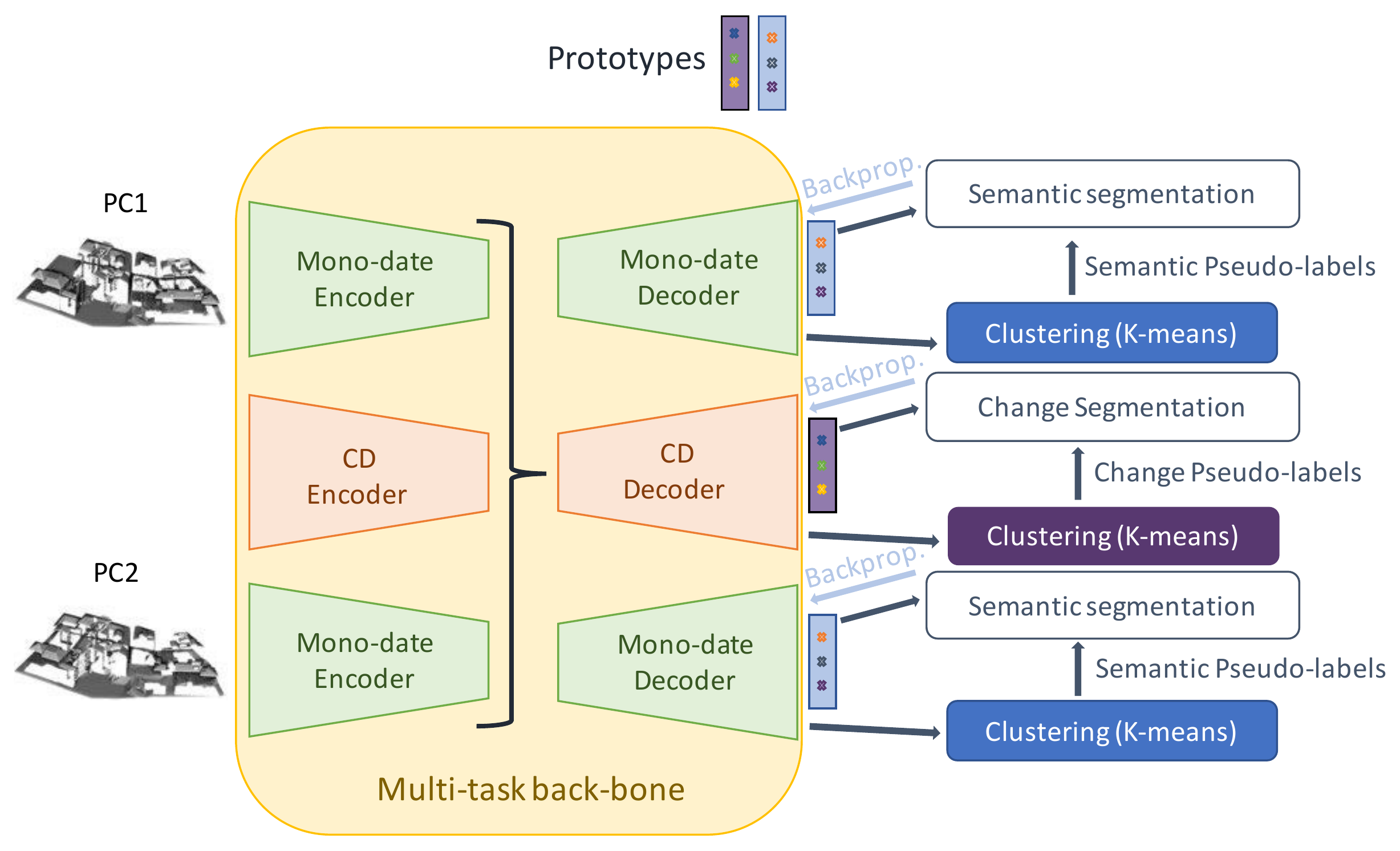}
    \caption[DC3DCD-V2 using multi-task learning.]{\textbf{DC3DCD-V2 using multi-task learning.} }
    \label{fig:DC3DCDV2}
\end{figure*}

\begin{table*}
    \centering
    \footnotesize
    \begin{tabular}{clcc|cc}
    \toprule
         &Method & i. f. & $y_{sim}$ & mAcc (\%)& mIoU$_{ch}$  (\%)\\
        \hline
        \multirow{2}{*}{\rotatebox{90}{Sup.}}
        &SKPConv \citep{degelis2023siamese}& & & 91.21 $\pm$ 0.68 & 80.12 $\pm$ 0.02 \\
         &EFSKPConv \citep{degelis2023change}&& & \textbf{94.23} $\pm$ 0.88 & \textbf{85.19} $\pm$ 0.24\\

          \hline
          \multirow{12}{*}{\rotatebox{90}{Weakly supervised}}

            &\ac{DC3DCD} EFSKPConv& && 52.30 $\pm$ 2.41 & 37.75 $\pm$ 2.11\\
            &DC3DCD-V2 EFSKPConv& & GT & 83.45 $\pm$ 2.22 & 73.51 $\pm$ 3.74\\
            &DC3DCD-V2 EFSKPConv&  &\ac{M3C2} \citep{lague2013accurate} & 54.01 $\pm$ 3.54 & 39.59 $\pm$ 4.18\\
            &DC3DCD-V2 EFSKPConv& & \ac{C2C} \citep{girardeau2005change} & 39.74 $\pm$ 1.84 & 25.57 $\pm$ 1.97\\
            &DC3DCD-V2 EFSKPConv&& Multi-task ($K_{seg.sem.}=4$) &34.31  $\pm$ 5.85 & 19.21 $\pm$ 5.81\\
            &DC3DCD-V2 EFSKPConv&& Multi-task ($K_{seg.sem.}=500$) & 47.62 $\pm$ 6.76 & 32.07 $\pm$ 6.15\\
            \cline{2-6}
            &\ac{DC3DCD} EFSKPConv &\checkmark & & 68.45 $\pm$ 1.10 & 57.06 $\pm$ 0.41\\
            &DC3DCD-V2 EFSKPConv& \checkmark &  GT & \textbf{89.04} $\pm$ 0.70 & \textbf{82.63} $\pm$ 0.73\\
            &DC3DCD-V2 EFSKPConv &\checkmark &  \ac{M3C2} \citep{lague2013accurate}& 58.80 $\pm$ 2.14 & 46.42 $\pm$ 2.45\\
            &DC3DCD-V2 EFSKPConv& \checkmark &  \ac{C2C} \citep{girardeau2005change}& 42.01 $\pm$ 0.67 & 28.04 $\pm$ 0.60\\
            &DC3DCD-V2 EFSKPConv &\checkmark &  Multi-task ($K_{seg.sem.}=4$)& 53.04 $\pm$ 8.22 & 38.90 $\pm$ 8.51\\
            &DC3DCD-V2 EFSKPConv& \checkmark &  Multi-task ($K_{seg.sem.}=500$)& 62.95 $\pm$ 1.81 & 50.14 $\pm$ 3.85\\
        \bottomrule
    \end{tabular}\\
    \caption[Quantitative evaluation of DC3DCD-V2 low density LiDAR dataset]{\textbf{Quantitative evaluation of DC3DCD-V2 on Urb3DCD-V2 low density \ac{LiDAR} dataset.} \textit{Top}: supervised methods. \textit{Middle}: Weakly supervised methods with our proposed \ac{DC3DCD} and DC3DCD-V2 with Encoder Fusion SiamKPConv architecture without the addition of 10 hand-crafted features as input to the network. \textit{Bottom}: Weakly supervised methods with our proposed \ac{DC3DCD} and DC3DCD-V2 with Encoder Fusion SiamKPConv architecture and with the addition of 10 hand-crafted features (i. f.) as input to the network. }
    \label{tab:ResLid05DC3DCDV2Appendix}
\end{table*}

\begin{table*}
    \centering
    \scriptsize
        \begin{tabular}{clcc|ccccccc}
        \toprule
        & &&& \multicolumn{7}{c}{Per class IoU (\%)}\\
        &Method & i. f. & $y_{sim}$  & Unchanged & New building & Demolition & New veg. & Veg. growth& Missing veg.& Mobile Object\\
        \hline
        \multirow{2}{*}{\rotatebox{90}{Sup.}}
        &SKPConv & &  & 95.82 $\pm$ 0.48 & 86.68 $\pm$ 0.47 & 78.66 $\pm$ 0.47 & 93.16 $\pm$ 0.27 & 65.17  $\pm$ 1.37 & 65.46  $\pm$ 0.93 & 91.55 $\pm$ 0.60\\
           &EFSKPConv & & & \textbf{97.47} $\pm$ 0.04 & \textbf{96.68} $\pm$ 0.30 & \textbf{82.29} $\pm$ 0.16 & \textbf{96.52} $\pm$ 0.03 & \textbf{67.76} $\pm$ 1.51 & \textbf{73.50} $\pm$ 0.81 & \textbf{94.37} $\pm$ 0.54\\

        \hline
        \multirow{16}{*}{\rotatebox{90}{Weakly supervised}}
         &\ac{DC3DCD}  & & &90.90 $\pm$ 0.79 & 64.06 $\pm$ 5.13 & 54.35 $\pm$ 3.84 & 58.14 $\pm$ 20.03 & 1.45 $\pm$ 2.05 & 0.94 $\pm$ 0.78 & 47.57 $\pm$ 2.58\\
          & DC3DCD-V2& & GT & 97.28 $\pm$ 0.10 & 93.66 $\pm$ 1.30 & 75.24 $\pm$ 4.34 & 83.78 $\pm$ 6.00 & 49.52 $\pm$ 7.00 & 53.60 $\pm$ 11.98 & 85.28 $\pm$ 3.54\\
          &DC3DCD-V2&  &\ac{M3C2} & 93.02 $\pm$ 0.03 & 75.05 $\pm$ 5.28 & 57.43 $\pm$ 2.03 & 69.02 $\pm$ 5.77 & 10.45 $\pm$ 8.60 & 4.47 $\pm$ 4.22 & 21.11 $\pm$ 10.22\\
            &DC3DCD-V2& & \ac{C2C} &90.73 $\pm$ 0.45 & 64.93 $\pm$ 3.79 & 56.52 $\pm$ 3.74 & 10.76 $\pm$ 5.80 & 0.41 $\pm$ 0.56 & 0.44 $\pm$ 0.51 & 20.37 $\pm$ 6.96\\
            &DC3DCD-V2&&\makecell{Multi-task \\ \tiny{($K_{seg.sem.}=4$)}} &87.15 $\pm$ 2.38 & 29.55 $\pm$ 16.82 & 39.72 $\pm$ 7.84 & 28.11 $\pm$ 2.95 & 0.00 $\pm$ 0.00 & 0.08 $\pm$ 0.09 & 17.78 $\pm$ 11.09\\
            &DC3DCD-V2&& \makecell{Multi-task \\ \tiny{($K_{seg.sem.}=500$)}} &88.56 $\pm$ 2.11 & 40.44 $\pm$ 12.79 & 45.16 $\pm$ 12.73 & 50.12 $\pm$ 6.53 & 4.03 $\pm$ 5.09 & 0.43 $\pm$ 1.19 & 51.93 $\pm$ 6.90\\
             \cline{2-11}

            &\ac{DC3DCD} & \checkmark & & 93.96 $\pm$ 0.11 & 79.26 $\pm$ 0.68 & 67.88 $\pm$ 0.49 & 75.34 $\pm$ 2.81 & 19.48 $\pm$ 4.00 & 20.29 $\pm$ 2.90 & 80.10 $\pm$ 3.16\\
             &DC3DCD-V2& \checkmark &  GT& \textbf{97.73} $\pm$ 0.05 & \textbf{94.50} $\pm$ 0.20 & \textbf{81.10} $\pm$ 0.38 & \textbf{92.22} $\pm$ 0.92 & \textbf{61.32} $\pm$ 2.39 & \textbf{73.39} $\pm$ 1.18 & \textbf{90.12} $\pm$ 1.30\\
            &DC3DCD-V2 &\checkmark &  \ac{M3C2}& 93.17 $\pm$ 0.13 & 77.48 $\pm$ 1.85 & 65.34 $\pm$ 0.55 & 76.96 $\pm$ 5.26 & 25.65 $\pm$ 4.94 & 0.31 $\pm$ 0.54 & 32.76 $\pm$ 4.15\\
            &DC3DCD-V2& \checkmark &  \ac{C2C}& 92.25 $\pm$ 0.15 & 70.01 $\pm$ 2.77 & 66.56 $\pm$ 1.12 & 27.11 $\pm$ 5.78 & 0.00 $\pm$ 0.00 & 4.59 $\pm$ 0.64 & 0.00 $\pm$ 0.00\\
            &DC3DCD-V2 &\checkmark &  \makecell{Multi-task \\ \tiny{($K_{seg.sem.}=4$)}}& 91.96 $\pm$ 1.10 & 69.01 $\pm$ 7.24 & 63.07 $\pm$ 1.49 & 42.76 $\pm$ 13.31 & 4.11 $\pm$ 6.71 & 17.23 $\pm$ 7.28 & 31.20 $\pm$ 25.63\\
            &DC3DCD-V2& \checkmark &  \makecell{Multi-task \\ \tiny{($K_{seg.sem.}=500$)}}& 93.68 $\pm$ 0.48 & 78.23 $\pm$ 2.94 & 66.02 $\pm$ 1.46 & 71.49 $\pm$ 2.49 & 0.00 $\pm$ 0.00 & 16.07 $\pm$ 4.77 & 69.06 $\pm$ 16.59\\
             \bottomrule
        \end{tabular}
    \caption[Per class quantitative evaluation of DC3DCD-V2 low density LiDAR dataset]{\textbf{Per class quantitative evaluation of DC3DCD-V2 on Urb3DCD-V2 low density \ac{LiDAR} dataset.} \textit{Top}: supervised methods. \textit{Middle}: Weakly supervised methods with our proposed \ac{DC3DCD} and DC3DCD-V2 with Encoder Fusion SiamKPConv architecture without the addition of 10 hand-crafted features as input to the network. \textit{Bottom}: Weakly supervised methods with our proposed \ac{DC3DCD} and DC3DCD-V2 with Encoder Fusion SiamKPConv architecture and with the addition of 10 hand-crafted features (i. f.) as input to the network.}
    \label{tab:DC3DCDResLid05IouV2Appendix}
    \end{table*}

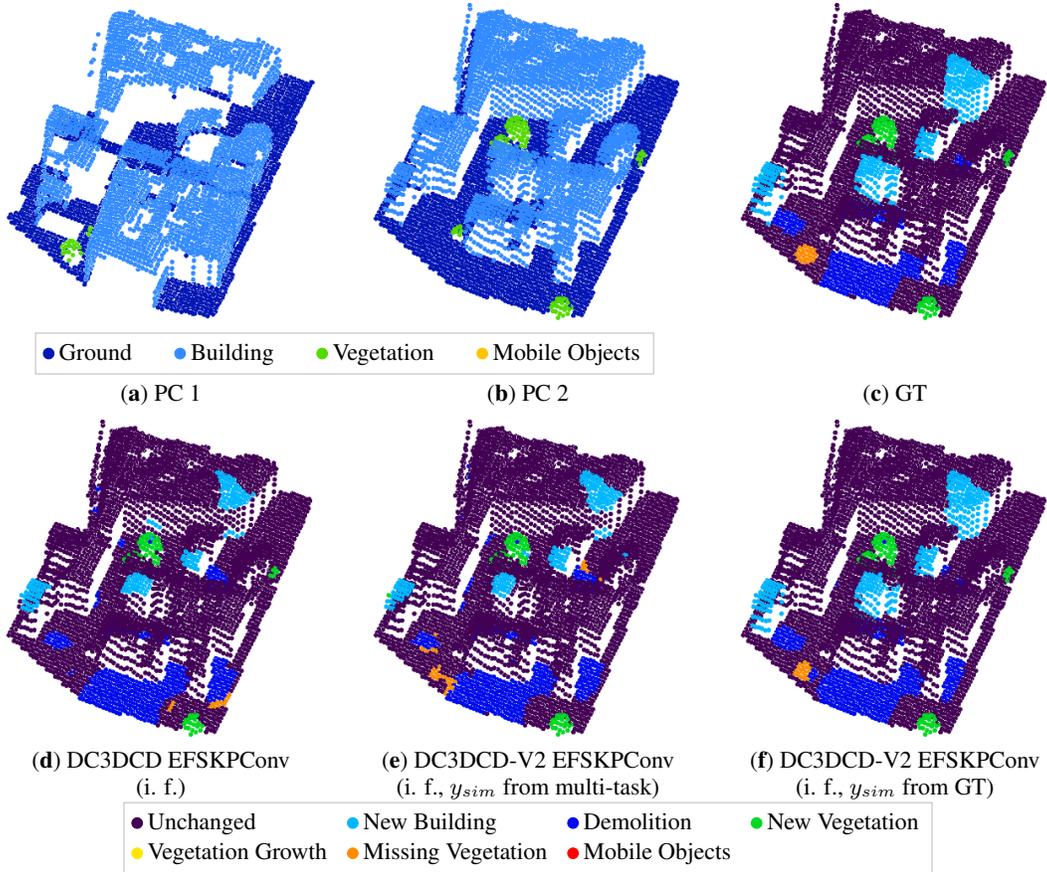
\begin{figure*}
    \centering
    \footnotesize
    \begin{tabular}{ccc}
        \includegraphics[width=0.27\textwidth]{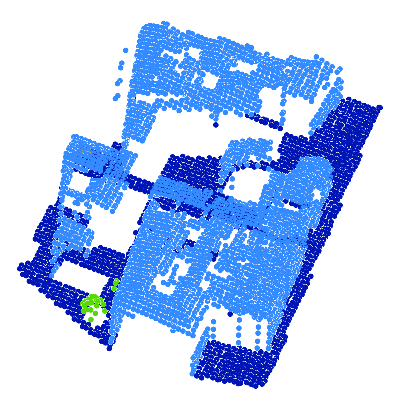} &
        \includegraphics[width=0.27\textwidth]{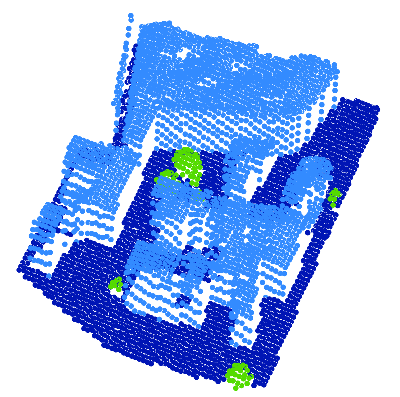} &
        \includegraphics[width=0.27\textwidth]{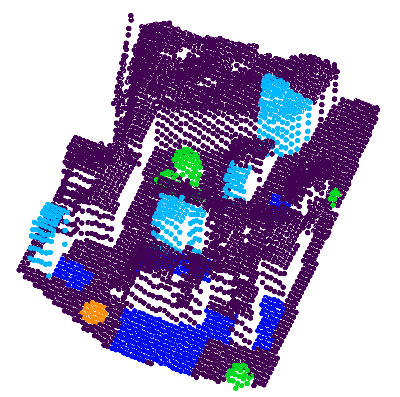}\\
        \multicolumn{2}{c}{        \begin{tikzpicture}
    		\begin{axis}[
            		xmin=1,
                    xmax=1.1,
                    ymin=1,
                    ymax=1.1,
                     hide axis,
    				width=0.5\textwidth ,
    				mark=circle,
    				scatter,
    				only marks,
    				legend cell align={left},
    				legend entries={Ground, Building, Vegetation, Mobile Objects},
    				legend style={draw=lightgray,at={(0,0)}, legend columns=4,/tikz/every even column/.append style={column sep=0.5cm}}]
    			\addplot[Ground] coordinates {(0,0)}; 
    			\addplot[Building] coordinates {(0,0)};
    			\addplot[Vegetation] coordinates {(0,0)};
    			\addplot[MO] coordinates {(0,0)};
    		\end{axis}
	    \end{tikzpicture}} & \\
        (\textbf{a}) PC 1 & (\textbf{b}) PC 2 & (\textbf{c}) GT \\

    \includegraphics[width=0.27\textwidth]{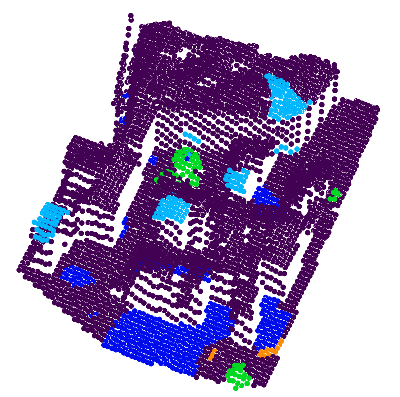}   &
        \includegraphics[width=0.27\textwidth]{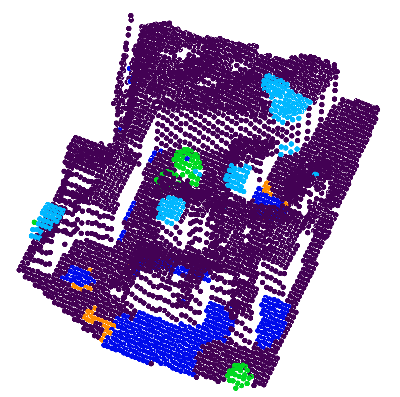}&
        \includegraphics[width=0.27\textwidth]{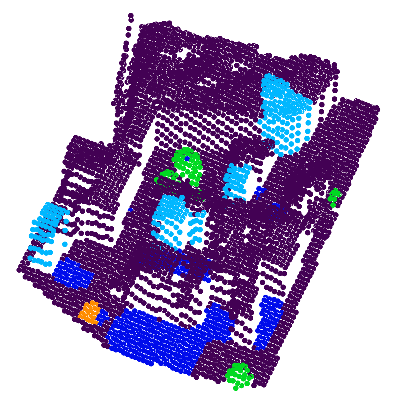}\\
        \makecell{(\textbf{d}) DC3DCD EFSKPConv \\(i. f.)} & \makecell{ (\textbf{e})  DC3DCD-V2  EFSKPConv \\
        (i. f., $y_{sim}$ from multi-task)}&\makecell{(\textbf{f}) DC3DCD-V2 EFSKPConv \\ (i. f., $y_{sim}$ from GT)  }\\

        \multicolumn{3}{c}{        \begin{tikzpicture}
    		\begin{axis}[
            		xmin=1,
                    xmax=2,
                    ymin=1,
                    ymax=2,
                     hide axis,
    				width=0.5\textwidth ,
    				mark=circle,
    				scatter,
    				only marks,
    				legend entries={Unchanged, New Building, Demolition, New Vegetation, Vegetation Growth, Missing Vegetation, Mobile Objects},
    				legend cell align={left},
    				legend style={draw=lightgray,at={(0,0)}, legend columns=4,/tikz/every even column/.append style={column sep=0.2cm}}]
    			\addplot[Unchanged] coordinates {(0,0)}; 
    			\addplot[NewBuild] coordinates {(0,0)};
    			\addplot[Demol] coordinates {(0,0)};
    			\addplot[VegeN] coordinates {(0,0)};
    			\addplot[VegeG] coordinates {(0,0)};
    			\addplot[VegeR] coordinates {(0,0)};
    			\addplot[MOch] coordinates {(0,0)};
    		\end{axis}
	    \end{tikzpicture}}\\
    \end{tabular}

    \caption[Qualitative assessment of DC3DCD-V2 method on Urb3DCD-V2 dataset]{\textbf{Visual change detection results on Urb3DCD-V2 low density \ac{LiDAR} sub-dataset:} (\textbf{a}-\textbf{b}) the two input point clouds; (\textbf{c}) ground truth (GT): simulated changes; (\textbf{d}) \ac{DC3DCD} with the \textit{Encoder Fusion SiamKPConv} architecture and 10 hand-crafted input features (i. f.) results; (\textbf{e}) \ac{DC3DCD}-V2  with the \textit{Encoder Fusion SiamKPConv} architecture, 10 hand-crafted input features results and the similarity $y_{sim}$ computed from the multi-task configuration($K_{mono-date}=500$); (\textbf{f}) \ac{DC3DCD}-V2 with the \textit{Encoder Fusion SiamKPConv} architecture, 10 hand-crafted input features results and the similarity $y_{sim}$ computed from the ground truth (GT).}
    \label{fig:DC3DCDresLid05-3}
\end{figure*}

\begin{figure*}
    \centering
  \begin{tabular}{ccc}
    \includegraphics[width=0.27\textwidth]{figure/vue3/GT.png}&
    \includegraphics[width=0.27\textwidth]{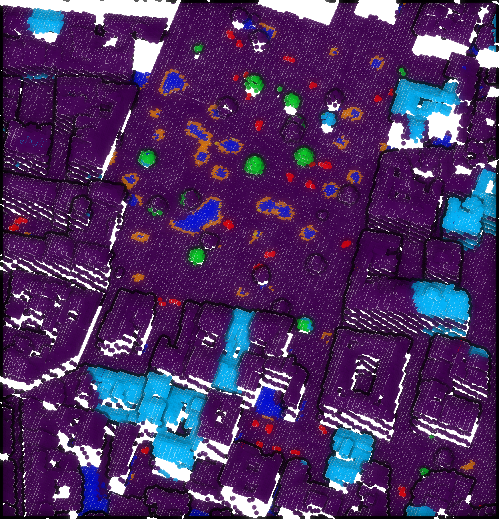}&
        \includegraphics[width=0.27\textwidth]{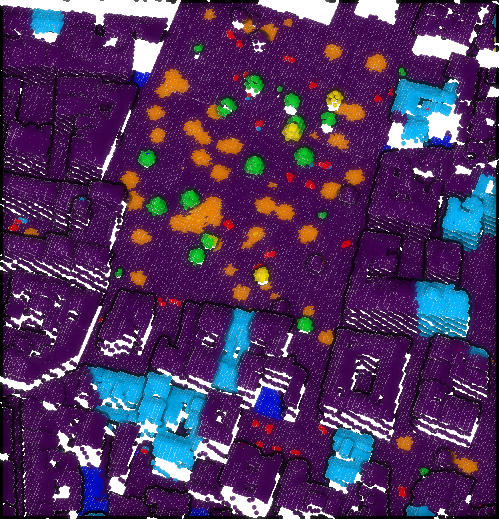}\\
        (\textbf{a}) Ground truth&
        \makecell{(\textbf{b}) DC3DCD-V2 EFSKPConv\\ (i. f., $y_{sim}$ from multi-task)} & \makecell{(\textbf{c}) DC3DCD-V2 EFSKPConv\\ (i. f., $y_{sim}$ from GT)}\\
        \multicolumn{3}{c}{        \begin{tikzpicture}
    		\begin{axis}[
            		xmin=1,
                    xmax=2,
                    ymin=1,
                    ymax=2,
                     hide axis,
    				width=0.5\textwidth ,
    				mark=circle,
    				scatter,
    				only marks,
    				legend entries={Unchanged, New Building, Demolition, New Vegetation, Vegetation Growth, Missing Vegetation, Mobile Objects},
    				legend cell align={left},
    				legend style={draw=lightgray,at={(0,0)}, legend columns=4,/tikz/every even column/.append style={column sep=0.2cm}}]
    			\addplot[Unchanged] coordinates {(0,0)}; 
    			\addplot[NewBuild] coordinates {(0,0)};
    			\addplot[Demol] coordinates {(0,0)};
    			\addplot[VegeN] coordinates {(0,0)};
    			\addplot[VegeG] coordinates {(0,0)};
    			\addplot[VegeR] coordinates {(0,0)};
    			\addplot[MOch] coordinates {(0,0)};
    		\end{axis}
	    \end{tikzpicture}}\\
    \end{tabular}
    \caption[Qualitative assessment of DC3DCD-V2 method on Urb3DCD-V2 dataset (area~2)]{\textbf{Visual change detection results on Urb3DCD-V2 low density \ac{LiDAR} sub-dataset (area~2):} (\textbf{a}) ground truth (GT): simulated changes; (\textbf{b}) \ac{DC3DCD}-V2 with the \textit{Encoder Fusion SiamKPConv} architecture, 10 hand-crafted input features (i. f.) and the similarity $y_{sim}$ computed from the multi-task configuration($K_{mono-date}=500$); (\textbf{c}) \ac{DC3DCD}-V2  with the \textit{Encoder Fusion SiamKPConv} architecture, 10 hand-crafted input features and the similarity $y_{sim}$ computed from the ground truth (GT). For comparison, one can refer to Figure~\ref{fig:DC3DCDresLid05-2} providing $k$-means and \ac{DC3DCD} results over the same area.}
    \label{fig:DC3DCDresLid05-2bis}
\end{figure*}


Another idea is to rely on multi-task learning \citep{vandenhende2021multi,zhang2021survey}: a multi-task framework based on \ac{DC3DCD} that jointly extracts mono-date features that can be used for similarity computation has been designed. As illustrated in Figure~\ref{fig:DC3DCDV2}, we added decoders for mono-date semantic segmentation to the back-bone architectures to also obtain semantic segmentation of \acp{PC}. Both \textit{Siamese KPConv} and  \textit{Encoder Fusion SiamKPConv} have encoders to extract mono-date features, therefore we just added a decoder taking as input these mono-date features instead of feature differences for \textit{Siamese KPConv} for example. Thereby, we used the same idea as before to train the network but with two separate clusterings, performed on output features of the change decoder on one side, as in previous experiments, and  on output features of mono-date decoders on the other side. This results in both change pseudo-labels and mono-date pseudo-labels which are used to modulate change encoder-decoder and mono-date encoders-decoders respectively. In practice, we shared trainable parameters between mono-date encoders and decoders. We trained first the semantic segmentation part and then apply a binary clustering (using $k$-means) on the nearest point mono-date features difference to obtain the similarity $y_{sim}$ used in the contrastive term of the change detection loss. Concerning, the number of pseudo-clusters for mono-date $K_{mono-date}$ , 4 and 500 have been tested (4 is the number of semantic segmentation classes in Urb3DCD-V2 dataset, 500 is considered as a sample large bound). While semantic segmentation scores, using the same user-guided mapping for mono-date semantic segmentation, are very promising (90.79\% of \ac{mIoU} on the 4 semantic segmentation classes of Urb3DCD-V2 with hand-crafted input features and $K_{mono-date}=500$), using the associated $y_{sim}$ still leads to unsatisfactory results (see Table~\ref{tab:ResLid05DC3DCDV2Appendix} and Table~\ref{tab:DC3DCDResLid05IouV2Appendix}). Indeed, when $K_{mono-date}$ is set to 500, we obtain 50.14\% of mIoU$_{ch}$ (with hand-crafted features) which is better than with distance-based methods (\ac{M3C2} \citep{lague2013accurate} or \ac{C2C} \citep{girardeau2005change}) but worse than without the contrastive part of the loss. Thereby, computing similarity from the nearest point mono-date features difference does not seem adapted. Two main reasons for this non-success can be advanced: i) the nearest point comparison is not optimal in occluded parts as well as in dense urban areas (which is the case for Urb3DCD datasets that are acquired on models of Lyon city center), and ii) there might be some side effects of the oversegmentation strategy. To explain, the method predicts 500 different mono-date semantic classes (far more than existing real classes). The oversegmentation occuring on the mono-date semantic segmentations leads to splitting each real semantic class into numerous semantic pseudo-classes. In other words, two points that belong to the same real semantic class will very likely be assigned to different semantic pseudo-classes (both pseudo-classes corresponding to the same real class). Since the change information is obtained by comparing the pseudo-classes and not the real ones, the similarity term $y_{sim}$ might be incorrect if the two pseudo-classes are far from each other in the feature space, and thus their difference is classified as changed during the binary clustering  step.
To counterbalance this oversegmentation issue, we tested using only 4 mono-date pseudo-clusters, but results are even worse, probably due to the fact that the DeepCluster strategy used to train a network requires a number of pseudo-clusters far greater than the number of real classes.
Thereby, we are facing here a conflicting situation, where the oversegmentation is needed to obtain an accurate mono-date semantic segmentation, but at the cost of errors in the identification of binary changes by comparing the mono-date semantic segmentations.

Qualitative assessment of these results is supported by   Figure~\ref{fig:DC3DCDresLid05-3} and Figure~\ref{fig:DC3DCDresLid05-2bis}. When the similarity comes from the binary change ground truth, visual results really lookalike the multi-change ground truth (Figure~\ref{fig:DC3DCDresLid05-3}f and Figure~\ref{fig:DC3DCDresLid05-2bis}c). Qualitative results of multi-task learning are quite encouraging. Surprisingly, borders of `missing vegetation' seems  well retrieved, but the center is still confused with demolition.

All these experiments aim at evaluating the potential of contrastive losses to improve our unsupervised results. Results with the similarity $y_{sim}$ issued from ground truth are very promising since they reach comparable results than the fully supervised networks. However, the method is highly dependent on the quality of this binary change annotation, and in case of mitigated binary annotation, it worsens \ac{DC3DCD} results. These first perspective experimentation are, to our opinion, encouraging to limit the annotation effort while preserving accurate results. Yet, the estimation of a precise $y_{sim}$ is still an open problem.

\section{Conclusion}\label{sec:ccl}

In this paper, we proposed an unsupervised learning method based on the DeepCluster principle to tackle multiclass change segmentation in raw \ac{3D} \acp{PC}. Following the unsupervised training, we propose a user-guided mapping of pseudo-clusters to real class in order to better to fit to the use case. Given the experiments on both synthetic and real dataset, we saw the importance of the choice of an appropriate architecture to extract valuable change-related features. Also, guiding the network using hand-crafted input features along with \ac{3D} points coordinates is advocated.

Using these recommended configuration, our proposed method, \acf{DC3DCD}, allows obtaining better results than a fully supervised traditional machine learning algorithm relying on hand-crafted features and to reach scores of fully supervised deep networks trained on 2.5D rasterization of \acp{PC}. We further proposed to improve \ac{DC3DCD} by introducing a contrastive loss leading to results comparable to those provided by fully supervised deep networks (89.04\% of mean of accuracy).

This last setting is very promising but was considered here in an ideal scenario where the similarity boolean used in the contrastive loss is faultless. The main challenge that remains to be solved is thus to be able to estimate the similarity term directly from the data. Another future work that would be of interest is to consider a multi-level  classification scenario, using a hierarchical clustering technique instead of a flat one. Finally, semi-automatic strategies to speed-up labeling the K clusters would also be of high benefit to ease deployment of our method.

\section*{Acknowledgements}
This research was funded by Magellium, Toulouse and the CNES, Toulouse. This work was granted access to the HPC resources of IDRIS under the allocation 2022-AD011011754R2 made by GENCI.

\begin{acronym}
    \acro{2D}{two-dimensional}
    \acro{3D}{three-dimensional}
    \acro{AHN}{Actueel Hoogtebestand Nederland}
    \acro{AHN-CD}{\ac{AHN} Change Detection}
    \acro{ALS}{Aerial Laser Scanning}
    \acro{C2C}{cloud-to-cloud}
    \acro{CNN}{Convolutional Neural Network}
    \acro{CO3D}{3D Optical Constellation}
    \acro{DC3DCD}{DeepCluster 3D Change Detection}
    \acro{DCVA}{Deep Change Vector Analysis}
    \acro{DSM}{Digital Surface Model}
    \acro{DTM}{Digital Terrain Model}
    \acro{FCN}{fully convolutional network}
    \acro{GCN}{Graph Convolution Network}
    \acro{GPU}{graphics processing unit}
    \acro{GT}{ground truth}
    \acro{H3D}{Hessigheim 3D}
    \acro{HD}{high density}
    \acro{IoU}{Intersection over Union}
    \acro{kNN}{k-Nearest Neighbors}
    \acro{KPConv}{Kernel Point Convolution}
    \acro{KP-FCNN}{Kernel Point -- Fully Convolutional Neural Network}
    \acro{LiDAR}{Light Detection And Ranging}
    \acro{M3C2}{Multi-Scale Model-to-Model Cloud Comparison}
    \acro{mAcc}{mean of accuracy}
    \acro{mIoU}{mean of \ac{IoU}}
    \acro{MLP}{Multi-Layer Perceptron}
    \acro{MLS}{Mobile Laser Scanning}
    \acro{NLL}{negative log-likelihood}
    \acro{NMI}{normalized mutual information}
    \acro{PC}{Point Cloud}
    \acro{PCA}{Principal Component Analysis}
    \acro{PDAL}{Point Data Abstraction Library}
    \acro{PIC}{power iteration clustering}
    \acro{RF}{Random Forest}
    \acro{RGB}{Red Green Blue}
    \acro{SAR}{Synthetic Aperture Radar}
    \acro{SGD}{Stochastic Gradient Descent}
    \acro{SiamGCN}{Siamese Graph Convolutional Network}
    \acro{SSL}{Self-Supervised Learning}
    \acro{TLS}{Terrestrial Laser Scanning}
    \acro{UAV}{unmanned aerial vehicle}
    \acro{Urb3DCD}{Urban 3D Change Detection}
    \acro{Urb3DCD-Cls}{Urban 3D Change Detection Classification}
\end{acronym}

\bibliographystyle{unsrtnat}
\bibliography{references}  






\end{document}